\documentclass{article}
\ifdefined\pdfminorversion\pdfminorversion=7\fi
\usepackage[T1]{fontenc}
\usepackage{iclr2027_conference,times}
\usepackage{amsmath,amssymb}
\usepackage{graphicx,booktabs}
\usepackage{caption}
\usepackage{placeins}
\usepackage{hyperref}
\usepackage{url}
\graphicspath{{figures/}}
\hypersetup{pdftitle={TSGate: Timestep-Aware Gated Attention in Diffusion Transformers},pdfauthor={}}

\iclrfinalcopy

\title{TSGate: Timestep-Aware Gated Attention\\for Diffusion Transformers}

\begin{document}
\raggedbottom
\maketitle

\begin{center}
    \vspace{-1.8cm}
    \large \textbf{Boyu Zhang$^{1,2,*}$, Yifan Liu$^{2,*}$, Shuxia Lin$^{2}$, \\ Qingjian Ni$^{2}$, Yinfei Xu$^{2}$, Xu Yang$^{2}$} \\
    \vspace{0.5em}
    \normalsize
    $^1$Alibaba Token Hub, Alibaba Group \quad $^2$Southeast University \\

    \vspace{0.8em}
\end{center}

\begingroup
\renewcommand{\thefootnote}{\fnsymbol{footnote}}
\footnotetext[1]{Contributed equally.}
\endgroup

\begin{abstract}
Diffusion Transformers (DiTs) have emerged as the dominant architecture for high-fidelity image and video generation. Recent DiT systems increasingly use structured prompts for training, improving caption quality and prompt adherence. However, their generation quality can degrade severely under out-of-domain (OOD) prompts, including the free-form descriptions supplied by users at inference time. Although LLM-based rewriting can convert these prompts into structured formats, it does not guarantee that the rewritten prompts align with the training distribution. Our analysis links this degradation to attention sinks and reduced early-step image-to-text attention and shows that sink suppression alone is insufficient to restore generation quality. Despite effective sink suppression, models trained with standard gated attention exhibit reduced early-step image-to-text attention and suboptimal generation quality. Based on these insights, we propose Timestep-Aware Gated Attention (TSGate), which injects a timestep-conditioned bias into the gate signal so that gating behavior adapts across denoising steps. Extensive experiments show that TSGate consistently outperforms both the baseline and standard gated attention across multiple benchmarks, improving the raw-prompt DPG score by 9.5\% over the baseline.

\end{abstract}

\section{Introduction}
\label{sec:intro}

Diffusion models have witnessed remarkable progress in visual content generation over the past few years~\citep{ho2020ddpm,rombach2022ldm,song2021scorebased}. Among the most impactful architectural shifts is the adoption of Transformer backbones in place of convolutional U-Nets, giving rise to the {Diffusion Transformer} (DiT) family~\citep{peebles2023dit,bao2023uvit}. DiTs now underpin state-of-the-art systems for both image generation (e.g., SD3~\citep{esser2024mmdit}, Flux~\citep{flux2024}, and PixArt-$\alpha$~\citep{chen2024pixart}) and video generation (e.g., Sora~\citep{brooks2024sora}, CogVideoX~\citep{yang2024cogvideox}, and Wan~\citep{wang2025wan}). A key enabler of their success is the joint attention paradigm introduced by MMDiT~\citep{esser2024mmdit}, which concatenates text and image tokens into a single sequence and lets them interact through shared self-attention layers. Concurrently, industrial practice~\citep{krea2026report,happyhorse26,seedance2026seedance20advancingvideo} has converged on training with structured prompts, which are detailed, template-based captions organized under explicit section headings. This approach yields faster convergence and lower training loss and has become a widely adopted standard for improving text--image alignment during training.

Yet the impressive generation quality observed under in-domain structured prompts can obscure a fragile dependence on the specific prompt format. At inference time, real users write natural-language descriptions that are free-form, unstructured, and often terse. These descriptions deviate significantly from the training distribution. Although prompt engineering is standard practice~\citep{seedance2026seedance20advancingvideo, 2026h3, betker2023dalle3, hao2023optimizing}, it cannot fully eliminate the underlying distribution shift. We find an overall decline in quality across prompt conditions ranging from mild perturbations of the structured template to full OOD inputs. This is a critical practical bottleneck; no deployment-ready DiT can expect users to craft perfectly structured prompts.

To investigate this degradation, we systematically analyze attention dynamics within the DiT during denoising. Two related phenomena emerge. First, we observe that DiT attention maps exhibit \emph{attention sinks}, characterized by a disproportionate concentration of attention mass on specific tokens, echoing discoveries in large language models~\citep{xiao2024streamingllm} and, more recently, in diffusion Transformers~\citep{wu2026attentionsink,li2026templatetokens}. Sinks are exacerbated under OOD prompts, for which the absence of familiar header tokens disrupts established attention patterns. Second, OOD prompts induce significantly less image-to-text attention than in-domain inputs during early denoising, when noisy image tokens must acquire global semantic structure from the text condition. This early-step attention deficit limits how much textual guidance image tokens receive when global semantics are being established, potentially compromising prompt fidelity in the final image.

\begin{figure}[t]
\centering
\includegraphics[width=\linewidth]{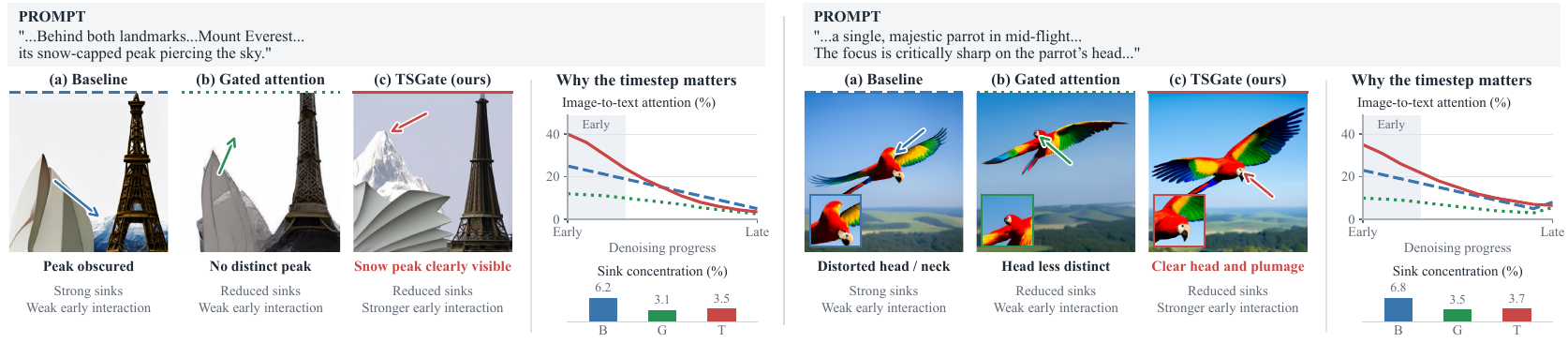}
\vspace{-0.2cm}
\caption{\textbf{Sink suppression is not enough; early text interaction matters.} For two out-of-domain prompts, TSGate renders the requested snow-capped peak and parrot's head more clearly than the baseline and standard gated attention. The schematic curves and bars illustrate the motivation for timestep-aware gating: reducing attention sinks must be accompanied by stronger text guidance during early denoising.}
\label{fig:intro-motivation}
\vspace{-0.0cm}
\end{figure}

A natural remedy is gated attention, which has been shown to effectively eliminate attention sinks and improve robustness in large language models~\citep{qiu2025gatedattention}. One would therefore expect it to address the sink problem in DiTs and consequently improve OOD generation quality. Despite substantially suppressing attention sinks, standard gated attention delivers suboptimal generation quality. Figure~\ref{fig:intro-motivation} illustrates this disconnect: reduced sinks coexist with weak early image-to-text interaction and unresolved errors in prompt-specified details. Sink suppression alone is therefore insufficient; the model must also preserve early text interaction. Although the input hidden states of the standard gate are timestep conditioned, the gate has no separately parameterized timestep offset. Because image tokens are dominated by noise in the early denoising phase, the content-dependent gate tends to suppress the image-to-text pathway. This finding exposes a fundamental mismatch: unlike autoregressive language models, diffusion models repeatedly process inputs whose statistics change dramatically with the noise level. Text--image interaction therefore requires a gate that is aware of the denoising stage.

Building on these insights, we propose \textbf{Timestep-Aware Gated Attention} (\textbf{TSGate}), a simple yet effective extension that explicitly conditions the gate on the current timestep. Concretely, we augment the standard gate with a timestep-conditioned bias. This design introduces only a small number of additional parameters per layer and is fully compatible with existing DiT architectures. Without explicit supervision of attention allocation, TSGate learns to maintain sink suppression while strengthening early image-to-text interaction, supporting semantic acquisition and more faithful rendering of prompt-specified details (Figure~\ref{fig:intro-motivation}).

Our main contributions are as follows:
\begin{itemize}
\item We identify the mechanistic root causes of out-of-domain prompt degradation in diffusion transformers: the emergence of attention sinks and the collapse of early-step image-to-text attention. By demonstrating that naive sink suppression is fundamentally insufficient, we establish a compelling need for stage-aware attention mechanisms.
\item We introduce TSGate, a lightweight, plug-and-play architectural enhancement that conditions attention gating on timestep information. This elegantly aligns gating behavior with the dynamic requirements of the denoising process while fully preserving token-specific content selection.
\item Extensive experiments demonstrate that TSGate significantly enhances out-of-domain generalization and robustness, consistently outperforming standard baselines. Furthermore, we show that a first-layer-only TSGate variant achieves these substantial benefits with negligible parameter overhead.
\end{itemize}

\section{Related Work}
\label{sec:related}
\vspace{-0.0cm}

\subsection{Diffusion Transformers}
\label{sec:related_dit}
\vspace{-0.0cm}

The transition from convolutional U-Net backbones~\citep{ho2020ddpm,rombach2022ldm} to pure Transformer architectures marks a defining shift in diffusion model design. \citet{peebles2023dit} introduced DiT, replacing the U-Net denoiser with a Vision Transformer equipped with Adaptive Layer Normalization (AdaLN) for timestep and class conditioning. Around the same time, \citet{bao2023uvit} proposed U-ViT, which retains long skip connections within a ViT backbone. Subsequent work has rapidly scaled DiTs: PixArt-$\alpha$~\citep{chen2024pixart} demonstrated efficient training through progressive resolution scaling, while Hunyuan-DiT~\citep{li2024hunyuandit} and Lumina-T2X~\citep{gao2024luminat2x} pushed multilingual and multimodal capabilities. A pivotal architectural advance is the joint attention (or MMDiT) paradigm introduced by SD3~\citep{esser2024mmdit}, in which text and image tokens are concatenated into a shared sequence for self-attention, superseding the earlier cross-attention design. Flux~\citep{flux2024} further streamlines this paradigm with flow-matching training. In the video domain, Sora~\citep{brooks2024sora}, CogVideoX~\citep{yang2024cogvideox}, and Wan~\citep{wang2025wan} extend DiTs to spatiotemporal modeling, while efficient variants such as DiT-MoE~\citep{fei2024ditmoe} explore mixture-of-experts scaling. In parallel, industrial practice has increasingly adopted structured, template-based captions for training~\citep{krea2026report,betker2023dalle3}, yet the implications of this practice for robustness, particularly vulnerability to OOD natural-language prompts at inference time, remain largely unexplored.

\vspace{-0.0cm}
\subsection{Attention Sinks in Transformers}
\vspace{-0.0cm}
\label{sec:related_sink}

The phenomenon of attention sinks, whereby a small number of tokens absorb a disproportionate share of attention mass regardless of their semantic relevance, was first systematically studied in autoregressive language models. \citet{xiao2024streamingllm} showed that the initial token in causal LLMs serves as an attention sink, and that preserving it is essential for stable streaming inference. In vision Transformers, \citet{darcet2024registers} demonstrated that dedicating explicit register tokens eliminates attention artifacts and improves representation quality. More recently, attention sinks have been investigated in diffusion Transformers. \citet{wu2026attentionsink} provided a causal analysis on sink behavior in DiTs. \citet{li2026templatetokens} discovered that certain text template tokens in the text encoder act as implicit semantic registers. \citet{su2026attentionsinksurvey} offered a survey of sink phenomena across Transformer families, and \citet{fu2026sinkaware} proposed sink-aware training objectives for mitigating the issue. Despite these advances, existing studies have largely characterized sinks in isolation; they have not provided an in-depth analysis of the sink problem in DiTs, nor have they explored what practical issues the sink problem actually causes for DiTs in real-world usage. Our work fills this gap by connecting sink analysis with early-step attention allocation and demonstrating that effective solutions must be timestep-aware.

\vspace{-0.0cm}
\subsection{Gated Attention Mechanisms}
\label{sec:related_gated}
\vspace{-0.0cm}

Gating mechanisms have a long history in sequence modeling, from LSTM~\citep{hochreiter1997lstm} gates to modern attention variants. \citet{qiu2025gatedattention} recently conducted a comprehensive study of gated attention in large language models, showing that introducing a learned sigmoid gate on attention outputs simultaneously adds beneficial nonlinearity, promotes sparsity, and eliminates attention sinks, yielding consistent improvements in LLM pre-training and downstream performance. The essence of gated attention lies in helping the network achieve a more optimal allocation of information. Several efficient-attention architectures incorporate related gating ideas: Gated Linear Attention (GLA)~\citep{yang2024gla} integrates data-dependent gating into linear-complexity attention, RetNet~\citep{sun2023retnet} employs exponential decay as an implicit gate, and RWKV~\citep{peng2023rwkv} uses channel-wise gating in a recurrent framework. In the diffusion domain, \citet{zhu2025dig} applied GLA to build an efficient diffusion backbone, and \citet{liu2026tgate} analyzed attention from a temporal perspective. \citet{chen2025dydit} proposed dynamic architectures that adapt computation per timestep, though without modifying the attention mechanism itself. Content-only output gating computes its logits from the current hidden representation, without a separate timestep offset in those logits. In DiTs, this representation can be timestep conditioned; TSGate complements that implicit path with an explicit bias branch, aligning content selection with the evolving requirements of denoising.
\vspace{-0.1cm}
\section{Method}
\vspace{-0.1cm}

\label{sec:method}
\begin{figure}[t]
\centering
\includegraphics[width=\linewidth]{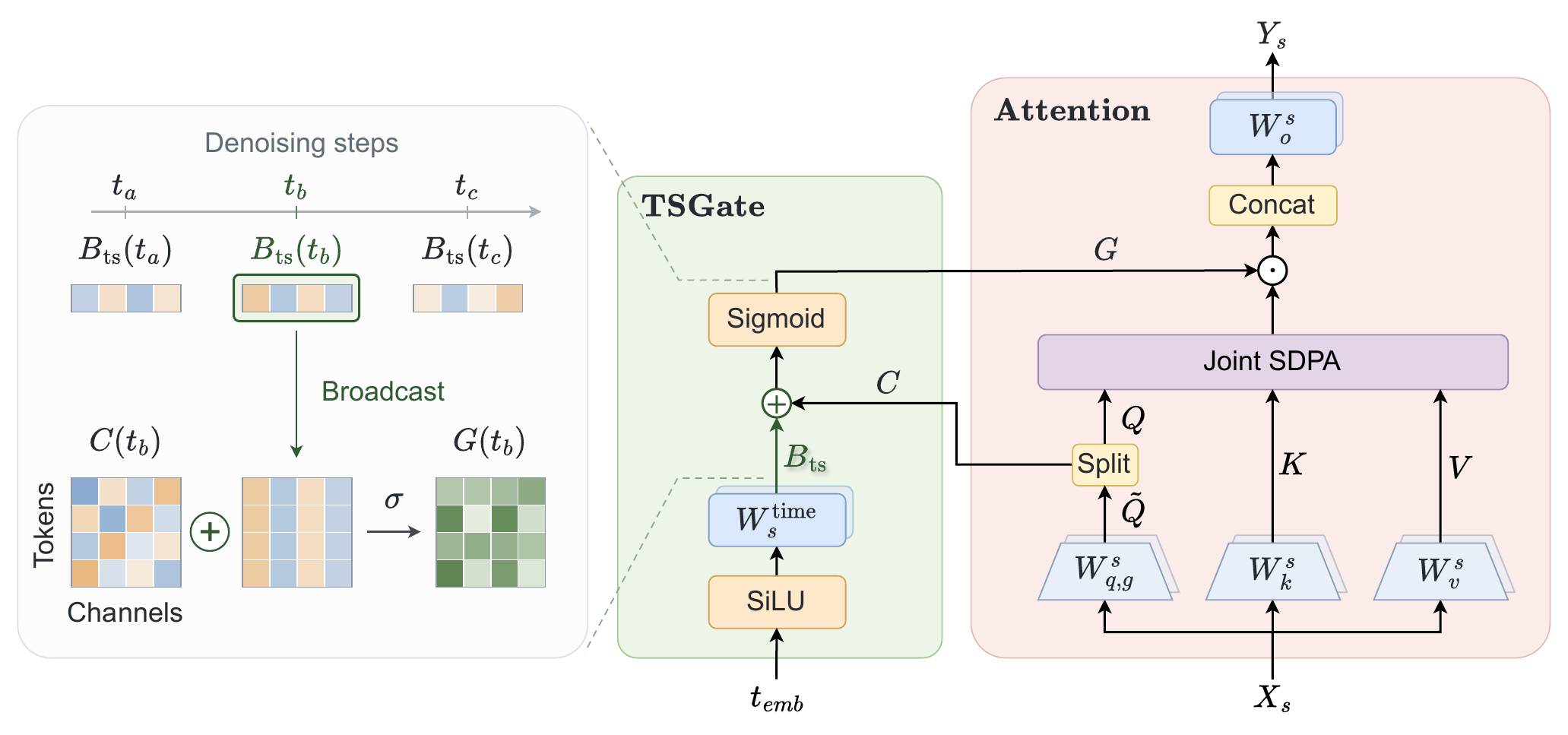}
\vspace{-0.6cm}
\caption{\textbf{Overview of TSGate.} At each denoising step $t$, a channel-wise time bias $B_{\mathrm{ts}}(t)$ is shared across tokens and added to the content logits $C$; a sigmoid produces the gate $G$. The gate scales the attention features before the output projection.}
\label{fig:architecture}
\vspace{-0.2cm}
\end{figure}

TSGate combines token-specific feature selection with explicit stage-dependent retention. We review content-only gating (standard gated attention~\citep{qiu2025gatedattention}), introduce the time branch, and explain how it complements the backbone's existing timestep conditioning.

\subsection{Preliminaries: Gated attention}
\label{sec:prelim}

Let $X\in\mathbb{R}^{N\times D}$ contain $N$ token features, with $H$ attention heads of width $d_h$ and $D=Hd_h$. For per-head queries, keys, and values, attention computes $A_h=\operatorname{softmax}(Q_hK_h^\top/\sqrt{d_h})V_h$ and concatenates the heads into $A\in\mathbb{R}^{N\times D}$. \emph{Gated attention} scales these aggregated features before the output projection. A widened query projection jointly produces queries $Q$ and content logits $C$:
\begin{equation}
  (Q,\,C)=\operatorname{split}\!\left(XW_{q,g}+b_{q,g}\right),\label{eq:gated-attention}
\end{equation}

\begin{equation}
Y=\left(A\odot\sigma\!\left(\operatorname{expand}(C)\right)\right)W_{o}+b_{o}.
\end{equation}

Here $W_{q,g}\in\mathbb{R}^{D\times(D+D_g)}$, $Q\in\mathbb{R}^{N\times D}$, and $C\in\mathbb{R}^{N\times D_g}$. For \emph{element-wise} gating ($D_g=D$), $\operatorname{expand}$ is the identity; for \emph{head-wise} gating ($D_g=H$), it repeats each head's logit across its $d_h$ channels. We use $\sigma$ for the element-wise sigmoid and $\odot$ for element-wise multiplication. We call this gate \emph{content-only}: its logits are computed from $X$.

\subsection{Timestep-aware gated attention}
\label{sec:computation}
We describe the default element-wise form for one sample and layer, omitting batch and layer indices. At timestep $t$, stream $s$ has input $X_s(t)\in\mathbb{R}^{N_s\times D}$ and aggregated attention features $A_s(t)\in\mathbb{R}^{N_s\times D}$. Following MMDiT~\citep{esser2024mmdit}, queries from each stream attend jointly to keys and values from all streams. We represent features as row vectors and use separate projection parameters for each layer and stream.

\paragraph{Content branch.}
As in content-only gating, the content branch in Figure~\ref{fig:architecture} uses a widened query projection to produce token-dependent logits:
\begin{equation}
(Q_s(t),C_s(t))=\operatorname{split}\!\left(X_s(t)W^s_{q,g}+b^s_{q,g}\right),
\label{eq:projection}
\end{equation}
where $W^s_{q,g}\in\mathbb{R}^{D\times2D}$ and $Q_s(t),C_s(t)\in\mathbb{R}^{N_s\times D}$. The split separates queries and logits within each head. Queries and keys follow the backbone's RMSNorm~\citep{zhang2019rmsnorm} and RoPE~\citep{su2021roformer} operations; content logits do not. Appendix~\ref{app:joint-computation} gives the tensor layouts and attention-path details.

\paragraph{Time branch.}
As shown in Figure~\ref{fig:architecture}, a separate projection maps the backbone timestep embedding $t_{\mathrm{emb}}(t)\in\mathbb{R}^{1\times d_t}$ to a channel-wise offset:
\begin{equation}
B_{\mathrm{ts},s}(t)=\operatorname{SiLU}\!\left(t_{\mathrm{emb}}(t)\right)W^{\mathrm{time}}_s+b^{\mathrm{time}}_s,
\label{eq:bias}
\end{equation}
Here SiLU~\citep{elfwing2017silu} is the activation function, $W^{\mathrm{time}}_s\in\mathbb{R}^{d_t\times D}$ and $b^{\mathrm{time}}_s\in\mathbb{R}^{1\times D}$. Unlike $C_s(t)$, the bias $B_{\mathrm{ts},s}(t)$ depends on the timestep embedding rather than on individual token features and is shared across the stream's tokens. Appendix~\ref{app:snr} explains why this separate, noise-free branch is useful when the content logits are dominated by noise at high timesteps.

We add the time bias to the content logits \emph{before} the sigmoid, then gate the aggregated attention features:
\begin{align}
G_s(t)&=\sigma\!\left(C_s(t)+\operatorname{broadcast}\!\left(B_{\mathrm{ts},s}(t)\right)\right),\label{eq:gate}\\
Y_s(t)&=\left(G_s(t)\odot A_s(t)\right)W^s_o+b^s_o.\label{eq:output}
\end{align}
Here $\operatorname{broadcast}$ repeats the time bias over $N_s$ tokens, so $G_s(t),Y_s(t)\in\mathbb{R}^{N_s\times D}$ and $W^s_o\in\mathbb{R}^{D\times D}$. For fixed content logits, increasing a channel's time bias raises its retention while preserving the ordering of gates across tokens; stage modulation thus remains content dependent. Appendix~\ref{app:properties} summarizes these local properties.

The gate acts on the value-aggregated attention features rather than on the softmax matrix. For a fixed query and channel, the same gate scales contributions from every key, so it neither renormalizes attention probabilities nor directly favors text keys over image keys within the current operation. Changes in the measured text-attention fraction instead emerge through training, subsequent layers, and denoising updates. This placement also differs from the backbone's adaLN-Zero conditioning~\citep{peebles2023dit}: its residual gate scales the projected attention branch, whereas $G_s(t)$ acts before $W_o^s$ and combines token-dependent content logits with an explicit timestep offset. Because pre-projection channel gating and post-projection residual scaling are generally not interchangeable, TSGate provides a distinct conditioning path. Appendices~\ref{app:output-gating} and~\ref{app:conditioning} give the formal details.

\vspace{-0.2cm}
\section{Experiments}
\label{sec:experiments}
\vspace{-0.1cm}

\subsection{Attention Sinks in Diffusion Transformers}
\label{sec:attention-sink}

\begin{figure}[t]
\centering
\includegraphics[width=\textwidth]{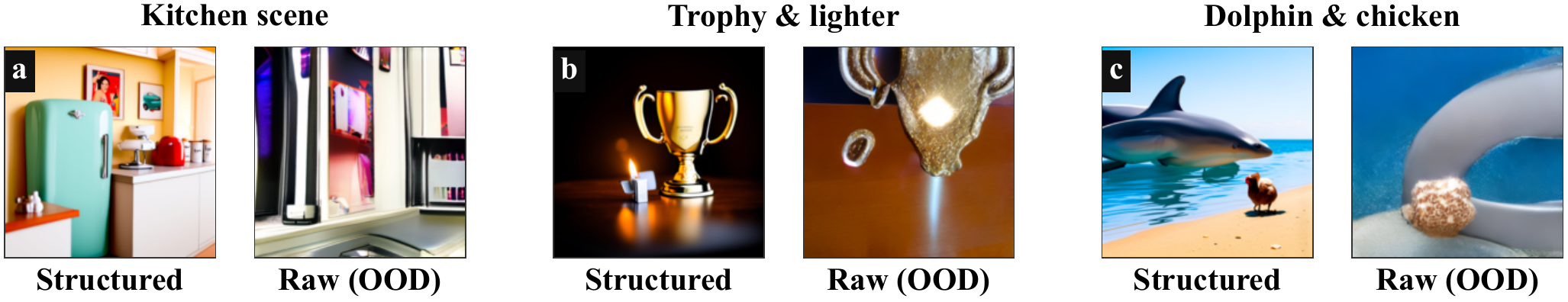}
\vspace{-0.5cm}
\caption{\textbf{Paired generations from in-domain structured prompts and OOD prompts.} Across three representative cases, departing from the training-time template substantially degrades object fidelity, compositionality, and scene coherence.}
\vspace{-0.2cm}
\label{fig:paired-cases}
\end{figure}

We begin by examining DiTs~\citep{peebles2023dit} trained on structured captions. Although these models generate high-quality images from in-domain prompts, their performance degrades severely on out-of-domain (OOD) inputs. Figure~\ref{fig:paired-cases} illustrates this contrast with three paired examples: generations from full-recaption structured prompts depict coherent scenes, whereas those from raw prompts omit requested objects, distort spatial relations, or lose semantic coherence. 

\begin{figure}[t]
\centering
\includegraphics[width=\textwidth]{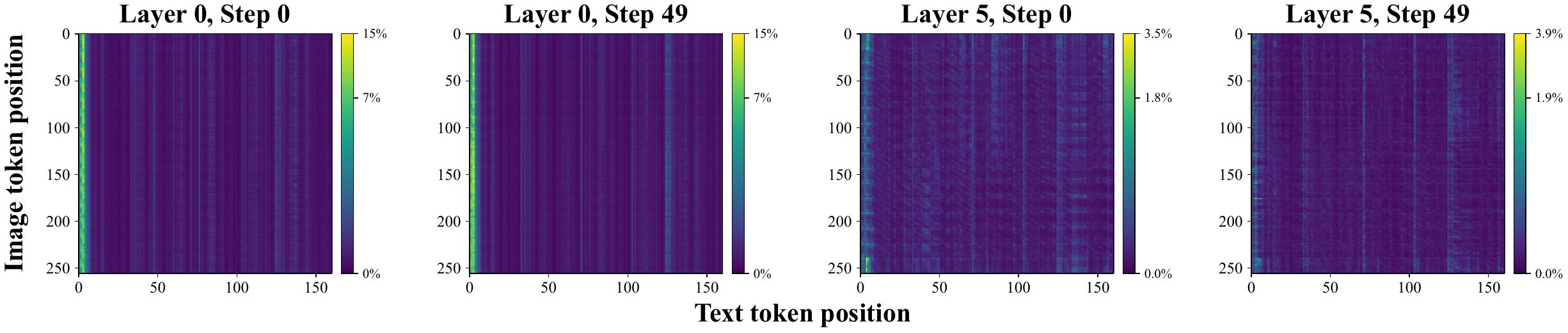}
\vspace{-0.2cm}
\caption{\textbf{Attention sinks persist across depth and denoising time.} Image-to-text attention maps for an in-domain example at two layers and two denoising steps. Bright vertical bands show attention concentrated on the same text tokens across image-token positions.}
\label{fig:attention-maps}
\end{figure}

Such failures are not surprising in isolation: free-form prompts are absent from the training distribution, so the model has no guarantee of preserving its behavior under this format shift. We therefore investigate the internal mechanism through which prompt format affects generation. Inspection of the joint-attention maps reveals a pronounced \emph{attention sink} phenomenon~\citep{xiao2024streamingllm,wu2026attentionsink}. During inference with in-domain structured prompts, a large fraction of attention is absorbed by a small set of fixed template tokens, particularly recurring section-heading markers. Figure~\ref{fig:attention-maps} visualizes image-to-text blocks from an in-domain example at two layers and two denoising steps, showing that these vertical sink bands persist across both depth and time.

Dataset-level statistics confirm that this concentration is substantial: recurring heading markers and separators such as "\textbackslash{}n" and "\#" receive more than $20\times$ the attention assigned to an average token. The model thus appears to rely on stable template markers with limited standalone semantic content as attention sinks. When these anchors disappear under OOD prompts, the learned attention organization is disrupted.

\begin{figure}[t]
\vspace{-0.0cm}
\centering
\includegraphics[width=\textwidth]{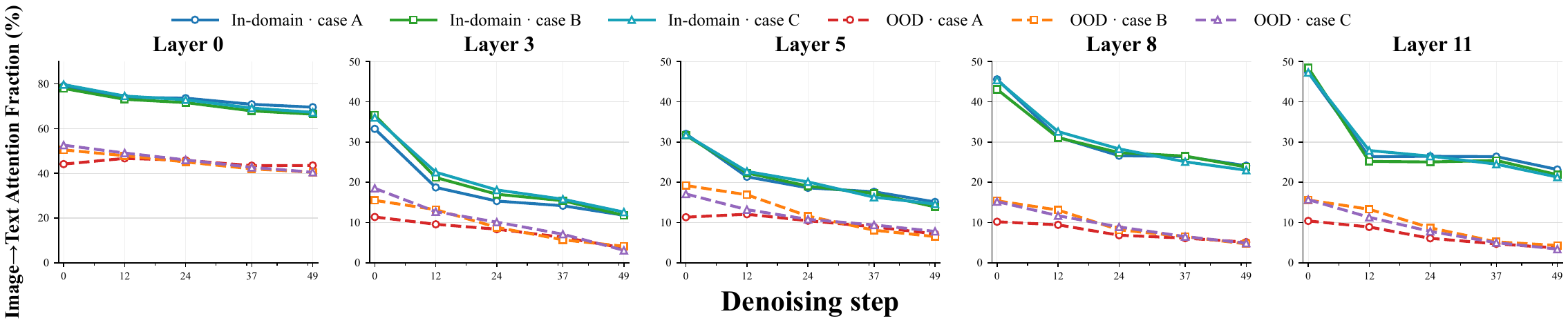}
\vspace{-0.2cm}
\caption{\textbf{Image-to-text attention across denoising for in-domain and OOD prompts.} Across five layers and three paired cases, OOD prompts (dashed curves) exhibit an early-step attention deficit relative to in-domain prompts (solid curves).}
\label{fig:ood-attention}
\vspace{-0.2cm}
\end{figure}

The timestep dimension exposes a second, complementary failure mode. As shown in Figure~\ref{fig:ood-attention}, image tokens allocate a large fraction of their attention to text early in denoising for in-domain prompts, then gradually reduce this interaction as visual structure emerges. OOD prompts lack this elevated initial level of text attention: their image-to-text attention fraction remains substantially below the in-domain trajectory precisely in the high-noise regime, when global semantics must be established. This deficit extends beyond the sink-dominated first layer: Layers 3, 5, 8, and 11 also lack the strong initial text interaction observed with in-domain prompts. The prompt shift therefore affects cross-modal information allocation throughout the backbone.

\vspace{-0.1cm}
\subsection{Text-to-Image Experiments}
\label{sec:t2i-experiment}
\vspace{-0.1cm}

The preceding analysis suggests that robust prompt generalization requires both controlling attention sinks and preserving early image-to-text interaction. We design TSGate to address these objectives jointly and evaluate it at scale on text-to-image generation. A natural alternative is standard gated attention~\citep{qiu2025gatedattention}, which is effective at suppressing sinks. However, it lacks a separately parameterized timestep offset. The results below confirm that directly transferring content-only gated attention to a DiT is suboptimal.

Introducing gated attention mid-training yields performance inferior to the baseline, since gate-controlled attention must be learned at the start~\citep{qiu2025gatedattention}. Therefore, we validate TSGate by training from scratch. Our training data comprise a curated 100M-image subset of LAION-5B~\citep{schuhmann2022laion5b}, re-captioned with Qwen2.5-VL 72B~\citep{bai2025qwen25vl} in a structured format detailed in Appendix~\ref{app:structured-prompt}. Further training and optimization details appear in Appendix~\ref{app:setup}.

\vspace{-0.1cm}
\subsubsection{Experimental Setup}
\label{sec:setup}
\vspace{-0.1cm}

We use a vanilla DiT as Baseline. Notably, to avoid the influence of differing parameter counts, Baseline should be scaled up to match the parameter of TSGate. Therefore, our primary experiments use a MoE architecture, which provides a simple way to align the total parameter count with TSGate by widening its shared experts. We use a backbone of 12-layer MoE DiT with 12 attention heads per layer and a head dimension of 128. Following the gate definitions in Section~\ref{sec:method}, the resulting configurations are Baseline (1.25B parameters), Gated attention (standard element-wise gated attention, 1.16B), and element-wise TSGate (1.25B).

We evaluate on DPG-Bench~\citep{hu2024ella} using both raw prompts and full-recaption prompts, GenEval~\citep{ghosh2023geneval} with 553 prompts, and T2I-CompBench~\citep{huang2023compbench} with 900 prompts. Unless stated otherwise, images are generated with a classifier-free guidance~\citep{ho2022classifierfree} scale of 5, a resolution of $320\times320$, and 50 denoising steps. Additional architecture, optimization, and evaluation details appear in Appendix~\ref{app:network-config}.

\subsubsection{Main Results}
\label{sec:main-results}

\begin{table}[t]
\caption{\textbf{TSGate improves generation quality across prompt formats and benchmarks.} We compare Baseline, Gated attention, and TSGate on DPG-Bench with raw and full-recaption prompts, and on GenEval and CompBench. Higher is better; DPG scores use a 0--100 scale, while the others use 0--1. Bold marks the best result among the three models shown; complete results and confidence intervals for all five variants appear in Appendix~\ref{app:quality}.}
\label{tab:main-results}
\label{tab:dpg}
\centering
\small
\begin{tabular*}{\linewidth}{@{\extracolsep{\fill}}lrrrr@{}}
\toprule
Model & DPG raw $\uparrow$ & DPG full recaption $\uparrow$ & GenEval $\uparrow$ & CompBench $\uparrow$ \\
\midrule
Baseline & 55.483 & 78.637 & 0.7861 & 0.5224 \\
Gated attention & 58.976 & 78.327 & 0.7862 & 0.5155 \\
TSGate & \textbf{60.749} & \textbf{79.355} & \textbf{0.8064} & \textbf{0.5236} \\
\bottomrule
\end{tabular*}
\end{table}

Table~\ref{tab:main-results} shows that TSGate consistently improves mean generation scores. On raw DPG-Bench prompts, TSGate scores 60.749, improving over Baseline by 5.266 points (9.5\%) and over Gated attention by 1.773 points. TSGate also obtains the highest full-recaption score, indicating that its benefit is not restricted to short OOD inputs. On compositional evaluation, TSGate improves GenEval from 0.7861 to 0.8064 and CompBench from 0.5224 to 0.5236 relative to Baseline. The TSGate$-$Gated-attention difference is positive on both external benchmarks (Appendix Table~\ref{tab:external}). The category breakdowns identify where these gains are concentrated (Appendix Tables~\ref{tab:geneval-categories} and~\ref{tab:comp-categories}).

\begin{figure}[t]
\centering
\includegraphics[width=1.0\textwidth]{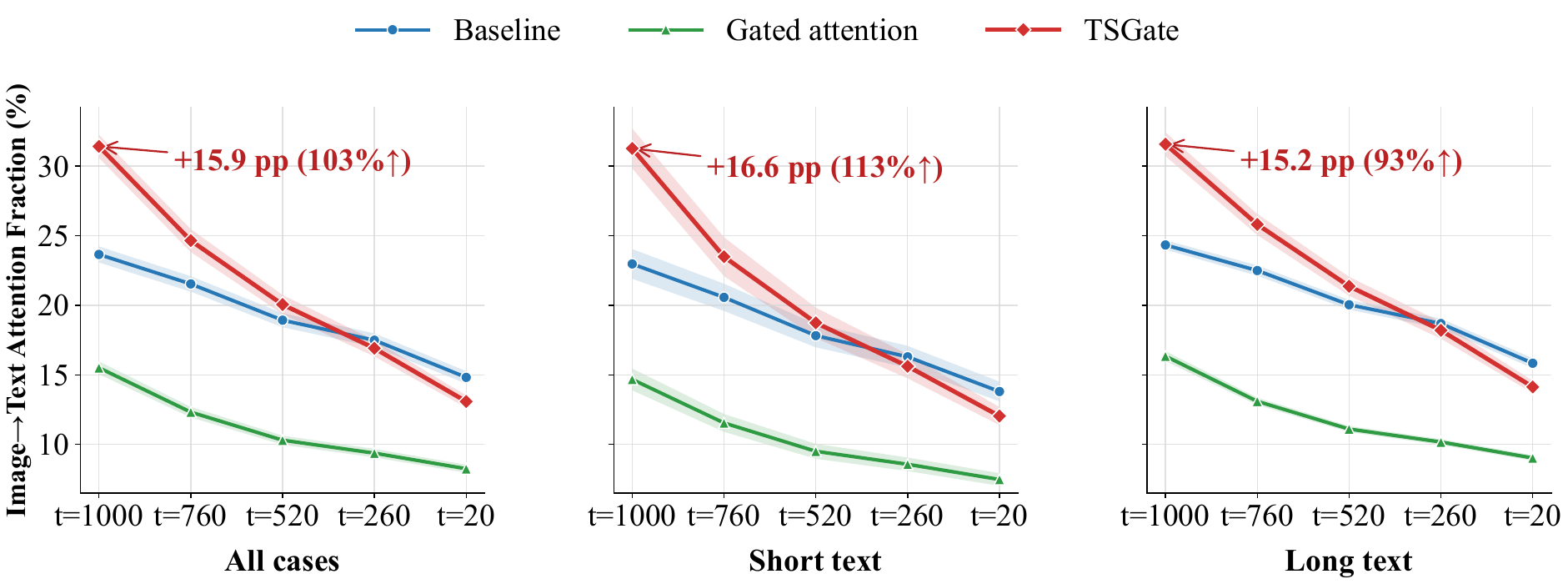}
\caption{\textbf{TSGate exhibits stronger early text attention and lower late text attention than Baseline.} Panels show Layer 5 image-to-text attention for all prompts, short prompts, and long prompts, from left to right. Shaded bands show 95\% confidence intervals. Red annotations report the early TSGate--Gated-attention gap in percentage points (pp), with the relative increase in parentheses.}
\vspace{-0.0cm}
\label{fig:tsgate-attention}
\end{figure}

\vspace{-0.2cm}
\paragraph{A learned shift from text acquisition to visual refinement.}
The attention trajectories in Figure~\ref{fig:tsgate-attention} help characterize this quality difference. TSGate exhibits stronger early image-to-text interaction: at $t=1000$, its attention fraction exceeds that of content-only gating by approximately 15.9 percentage points (103\% relative), and that of Baseline by 7.8 points. The early TSGate$-$Gated-attention gain appears in both equal-count length groups: 16.6 points for short prompts and 15.2 for long prompts. Thus, the aggregate increase is not driven solely by one prompt-length group.

More importantly, TSGate does not simply increase text attention throughout denoising. In all three panels, the red TSGate mean curve starts above the blue Baseline, crosses below it between the displayed $t=520$ and $t=260$ positions, and remains lower at $t=20$. This shared crossover indicates stage-dependent reallocation: image tokens attend more to text early, while their reduced late-stage text fraction is consistent with greater reliance on image--image interactions relative to Baseline. This pattern admits an intuitive coarse-to-fine interpretation of joint text--image interaction during denoising: noisy image tokens initially consult text to establish object identities and global layout; as visual structure emerges, interactions among image tokens become more useful for refining local appearance and detail. As noted in SD3~\citep{esser2024mmdit}, generation progresses differently across denoising stages: early steps establish coarse structure, whereas later steps refine details.

Neither a crossover nor a prescribed early/late attention schedule is built into TSGate. Moreover, its output gate does not directly renormalize text versus image probabilities (Appendix~\ref{app:output-gating}). The observed pattern therefore reflects an emergent stage-dependent allocation in the learned network, rather than a manually imposed switching rule. 

Indeed, measured first-layer sink suppression follows Gated attention $>$ TSGate $>$ Baseline (Appendix~\ref{app:controls}), whereas raw-prompt DPG quality follows TSGate $>$ Gated attention $>$ Baseline. Sink suppression alone therefore does not explain OOD robustness in DiTs. Content-only gating has the lowest text-attention fraction throughout Figure~\ref{fig:tsgate-attention}, including the early semantic-acquisition stage. TSGate instead combines reduced sink concentration with stronger early text interaction and a later shift away from text relative to Baseline. The useful objective is thus stage-appropriate allocation, not minimizing either sink concentration or text attention in isolation.

\vspace{-0.1cm}
\paragraph{Progressive prompt degradation.}
To isolate robustness to prompt format and descriptive context, we progressively move from the training distribution to raw benchmark inputs using six conditions. \emph{In domain} retains the complete structured prompt with section headings. \emph{Shuffle section} reorders the section headings while preserving the descriptive content, thereby disrupting its organization. \emph{W/o section} removes all heading markers (e.g., \texttt{\#Summary} and \texttt{\#Setting}) but retains the prose. \emph{W/o style} removes the mood and visual-style sections. \emph{Only subject} retains only the subject-description section and discards all other sections. Finally, \emph{OOD} replaces the structured description with the original short prompt from DPG-Bench, fully departing from the training-time format.

\begin{figure}[t]
\centering
\includegraphics[width=\textwidth]{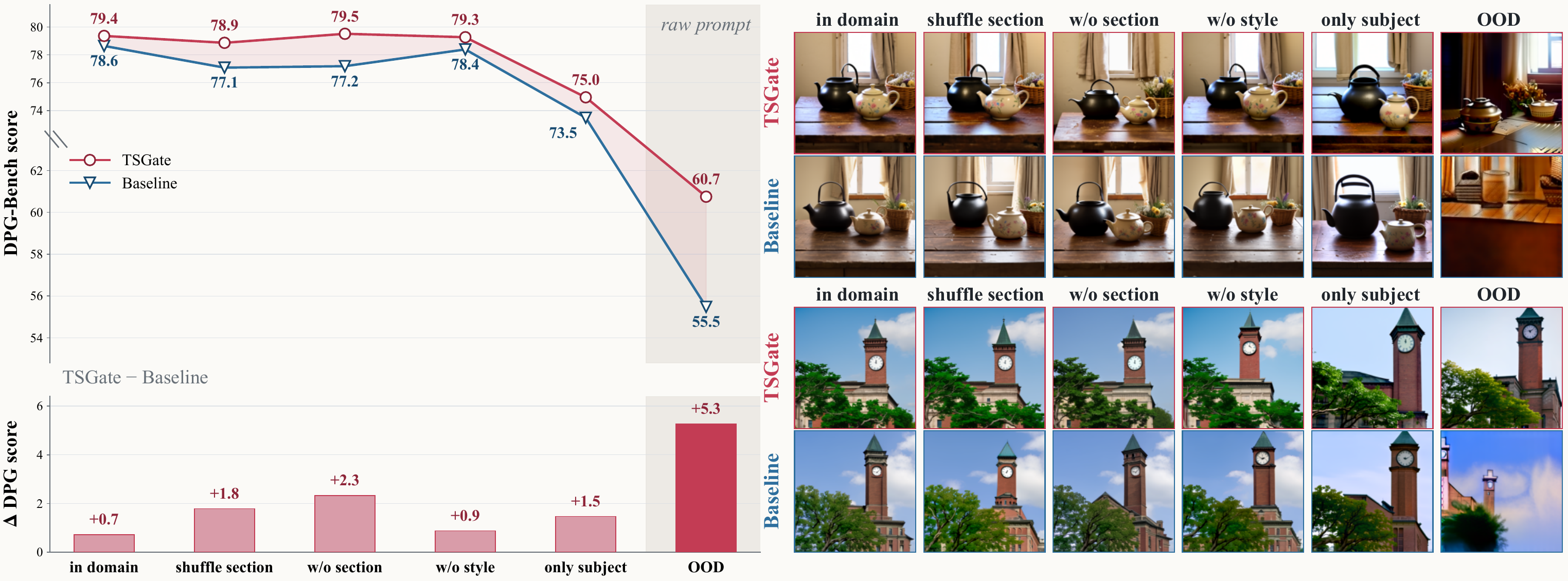}
\caption{\textbf{TSGate is more robust as prompt structure is removed.} Left: DPG-Bench scores and the TSGate--Baseline gap across six prompt conditions, from the structured training format to the raw OOD prompt. Right: examples under the same conditions, with TSGate above Baseline.}
\label{fig:prompt-degradation}
\vspace{-0.0cm}
\end{figure}

Figure~\ref{fig:prompt-degradation} shows that both models decline overall as structure and descriptive context are removed. TSGate remains better throughout, and its advantage is largest under the strongest shift: using unrounded scores, the gap grows from 0.7 points on in-domain prompts to 5.3 points on OOD prompts.

The intermediate conditions distinguish sensitivity to template organization from reliance on descriptive content. Shuffling sections preserves scene information yet lowers Baseline's score from 78.6 to 77.1, suggesting sensitivity to ordering beyond the facts supplied. Removing headings leaves TSGate near its in-domain score (79.5 versus 79.4), while Baseline's score falls to 77.2. This is consistent with TSGate making less use of template markers as indispensable anchors. When only the subject description remains, and when it is replaced by the raw prompt, TSGate still performs substantially better than Baseline. Further discussion is provided in Appendix~\ref{app:representations}.

The qualitative examples reinforce this trend. Under OOD prompts, Baseline exhibits pronounced compositional and color distortions in both the teapot and clock-tower scenes, whereas TSGate preserves recognizable subjects and produces substantially more coherent compositions.

\subsubsection{TSGate Family}
\label{sec:tsgate-family}

We further explore TSGate across gating configurations and backbone architectures, considering head-wise modulation, a dense backbone, and a first-layer time branch.

The head-wise variant, TSGate-head, replaces element-wise gates with one shared gate value per attention head in each stream and layer. Both the content and time branches use this coarser granularity. On raw DPG-Bench, TSGate-head scores 55.069, below element-wise TSGate. Appendix Tables~\ref{tab:moe-network-config} and~\ref{tab:dpg-moe} provide the configurations and complete results.

Experiments on a dense backbone further test whether TSGate's benefits extend beyond MoE architectures (Appendix~\ref{app:dense}). On raw prompts, TSGate improves DPG from 41.619 to 64.159, a gain of 22.540 points. Under full recaption, the mean increases from 81.8 to 82.3. These results support the applicability of TSGate beyond MoE, particularly for raw prompts. Note that the dense comparison is not parameter-matched. Architecture details appear in Appendix Table~\ref{tab:dense-network-config}.

Finally, we examine whether the timestep branch is needed in every layer. The sink score measurements in Appendix Figure~\ref{fig:attention} show the strongest concentration at the first transformer layer among the sampled layers. We also observe more pronounced cross-modal attention allocation in the first layer than in the other layers. This motivates TSGate-L0, which retains content gates in all layers but applies the timestep branch only at Layer 0. TSGate-L0 reaches 60.4 on raw DPG-Bench, compared with 60.7 for all-layer TSGate, retaining 93.8\% of its improvement over Baseline. Meanwhile, the time-branch parameter count drops from 85.0M to 7.1M (Appendix Table~\ref{tab:moe-network-config}). Thus, restricting the time branch to the first layer preserves most of the observed gain with fewer additional parameters.

\subsubsection{Time-bias interventions}
\label{sec:ablation}

To test the importance of temporal alignment, we reverse the learned time-bias schedule while keeping the checkpoint and backbone timestep conditioning fixed. This intervention reduces the raw DPG-Bench score by 2.6 points. Reversal preserves the learned bias values and their schedule average, changing only their assignment to denoising stages. The resulting degradation supports the functional importance of temporal alignment beyond the presence of an additive bias. Full intervention definitions and results are provided in Appendix~\ref{app:controls}.

\vspace{-0.0cm}
\section{Conclusion}
\label{sec:main-end}
\vspace{-0.0cm}

We investigate why Diffusion Transformers~\citep{peebles2023dit} trained on structured prompts degrade when given free-form user prompts. Our analysis links this fragility to attention sinks on template markers and reduced early-step image-to-text attention during high-noise denoising. Standard gated attention~\citep{qiu2025gatedattention} suppresses sinks but exhibits weaker early text interaction, showing that sink suppression alone is insufficient to restore generation quality. To address this mismatch, we propose TSGate, Timestep-Aware Gated Attention, a lightweight, plug-and-play modification that adds a timestep-conditioned bias to the gate signal. TSGate jointly reduces attention sinks and restores early text guidance, spontaneously learning a coarse-to-fine shift from semantic acquisition to visual refinement. It improves the raw-prompt DPG-Bench~\citep{hu2024ella} score by 9.5\% over the baseline while also achieving gains on GenEval~\citep{ghosh2023geneval} and CompBench~\citep{huang2023compbench}. Moreover, applying the timestep branch only to the first layer retains 93.8\% of the improvement with minimal parameter overhead, offering an efficient path toward improving DiT performance.

While our evaluation centers on text-to-image generation, extending TSGate to broader multimodal tasks (e.g., text-to-audio and text-to-video generation) remains to be explored. Moreover, visual tasks such as image editing require more nuanced spatial relationships between tokens than those typically modeled by LLMs. Combining TSGate with spatial positional encodings could thus be a worthwhile direction to explore.

\bibliography{references}
\bibliographystyle{iclr2027_conference}
\clearpage
\appendix

\section{Gate implementation and conditioning}
\label{app:variants}
\label{sec:variants}
\subsection{Granularity, layer scope, and initialization}
\paragraph{Gate granularity.}
Section~\ref{sec:computation} gives the default element-wise form of gated attention~\citep{qiu2025gatedattention}. More generally, let $D_g=D$ for element-wise gating and $D_g=H$ for head-wise gating. The content projection has $W^s_{q,g}\in\mathbb{R}^{D\times(D+D_g)}$ and $b^s_{q,g}\in\mathbb{R}^{1\times(D+D_g)}$, yielding $Q_s(t)\in\mathbb{R}^{N_s\times D}$ and $C_s(t)\in\mathbb{R}^{N_s\times D_g}$. The time projection has $W^{\mathrm{time}}_s\in\mathbb{R}^{d_t\times D_g}$ and $b^{\mathrm{time}}_s\in\mathbb{R}^{1\times D_g}$. The two forms share the computation
\begin{equation}
G_s(t)=\sigma\!\left(\operatorname{expand}\!\left(C_s(t)+\operatorname{broadcast}\!\left(B_{\mathrm{ts},s}(t)\right)\right)\right).
\label{eq:gate-granularity}
\end{equation}
Here $\operatorname{broadcast}$ repeats the time bias along the token dimension, and $\operatorname{expand}$ maps the $D_g$ gate channels to $D$ attention channels as in Equation~\eqref{eq:gated-attention}. Thus $G_s(t)\in\mathbb{R}^{N_s\times D}$ in either case, and Equation~\eqref{eq:output} is unchanged. Head-wise gating shares one sigmoid value across each head's $d_h$ channels; it changes the granularity of both branches.

\paragraph{Layer scope and evaluated configurations.}
We evaluate five MoE configurations (Table~\ref{tab:moe-network-config}). Baseline has no output gate. Gated attention gates content element-wise in all 12 layers. TSGate and TSGate-L0 retain this content gating, adding time biases in all layers and Layer 0, respectively. TSGate-head gates both branches head-wise in all layers; TSGate versus TSGate-head thus compares both branches' granularity. Dense experiments compare a no-output-gate baseline with full TSGate on a 24-layer backbone.

For layers without a time branch, Equation~\eqref{eq:gate-granularity} uses $B_{\mathrm{ts},s}(t)=0$ and no time-projection parameters. The first-layer configuration therefore retains content-only gating in Layers 1--11; it does not remove their output gates.

\paragraph{Initialization.}
For the MoE configurations, the content-logit entries of $b^s_{q,g}$ are initialized to 4, while $W^{\mathrm{time}}_s$ and $b^{\mathrm{time}}_s$ are initialized to zero. Consequently, $B_{\mathrm{ts},s}(t)=0$ at initialization, and the gate is exactly the corresponding content-only gate for the same content-branch parameters. If the token-dependent contribution to a logit is zero, its initial gate value is $\sigma(4)\approx0.982$; in general, that value also depends on the projected token features. This initialization favors retention but does not make the gated model exactly equivalent to an ungated baseline.

\subsection{Joint attention layout}
\label{app:joint-computation}
For element-wise gating, the projection in Equation~\eqref{eq:projection} is reshaped to $N_s\times H\times2d_h$ and split within each head into queries and content logits. Head-wise gating instead uses $N_s\times H\times(d_h+1)$, with one content logit per head. Keys and values use independent affine projections. Queries and keys undergo per-head RMSNorm~\citep{zhang2019rmsnorm} followed by RoPE~\citep{su2021roformer}; content logits undergo neither operation. Each stream's queries attend to the concatenated keys and values of all streams, following MMDiT~\citep{esser2024mmdit}, using the standard softmax attention computation~\citep{vaswani2017attention} in Section~\ref{sec:prelim}. Head concatenation gives $A_s(t)$, which is gated before the output projection. Gating before head concatenation is equivalent when the head--channel layout is preserved.

\subsection{Shared bias and content selection}
\label{app:properties}
For a fixed layer, stream, channel, and step, write $g_i(b)=\sigma(c_i+b)$. The sigmoid is strictly increasing, so a shared bias preserves the ordering of the given content logits across tokens. Also, $g_i(b)>1/2$ exactly when $c_i>-b$, and $\partial g_i/\partial b=g_i(1-g_i)\leq1/4$. Thus, a common offset changes retention most strongly for gates near $1/2$, while saturated gates respond less. These are local properties at fixed content logits; they do not require a monotone timestep schedule or guarantee an increase in text attention.

\subsection{Output gating does not renormalize attention probabilities}
\label{app:output-gating}
Let $P^h_{ij}$ be the pre-gate softmax probability from query $i$ to key $j$ in head $h$. For channel $c$, output gating gives
\begin{equation}
\widetilde A^h_{s,i,c}
=G^h_{s,i,c}A^h_{s,i,c}
=\sum_j G^h_{s,i,c}P^h_{ij}V^h_{j,c}.
\label{eq:gated-value-aggregation}
\end{equation}
The gate is shared across keys for a fixed query and channel, so these coefficients sum to $G^h_{s,i,c}$ rather than one. They do not define a new softmax distribution or selectively rescale text keys relative to image keys. Our attention metrics use the head-averaged $P$ (Equation~\eqref{eq:attention-metrics}). Output gating affects subsequent layers and denoising updates, and training can change the learned projections; it does not alter probabilities already computed in the same operation.

Using the head-averaged, pre-gate probability $P_{ij}$ from query $i$ to key $j$, we define the reported attention metrics as follows, where $\mathcal I$ is the image-query set, $\mathcal J$ is the text-key set, and $\mathcal K$ is the set of all keys:
\begin{equation}
\begin{aligned}
\mathrm{Image\text{-}to\text{-}text\ attention\ fraction}&=\frac{1}{|\mathcal I|}\sum_{i\in\mathcal I}\sum_{j\in\mathcal J}P_{ij},\\
\mathrm{Peak\ attention\ mass}&=\max_{j\in\mathcal K}\frac{1}{|\mathcal I|}\sum_{i\in\mathcal I}P_{ij}.
\end{aligned}
\label{eq:attention-metrics}
\end{equation}
Both statistics use image queries and take values in $[0,1]$. Peak attention mass searches all key modalities, whereas the image-to-text attention fraction sums over text keys. The image-to-text attention plots in the main text report this fraction as a percentage, multiplying it by 100.

\subsection{Relationship to adaLN-Zero and residual gating}
\label{app:conditioning}
Following the adaLN-Zero formulation of DiT~\citep{peebles2023dit}, let $U_s$ be the residual-stream input and $\gamma_s(t)$, $\beta_s(t)$, and $\alpha_s(t)$ the backbone's channel-wise scale, shift, and residual gate. Omitting the MLP branch and token broadcasting,
\begin{equation}
\begin{aligned}
X_s(t)&=\bigl(1+\gamma_s(t)\bigr)\odot\operatorname{LN}(U_s)+\beta_s(t),\\
Y_s(t)&=\bigl(G_s(t)\odot A_s(t)\bigr)W_o^s+b_o^s,\\
U_s^+&=U_s+\alpha_s(t)\odot Y_s(t).
\end{aligned}
\label{eq:conditioning-paths}
\end{equation}
The content logits already depend on $t$ through $X_s(t)$. TSGate adds a separately parameterized offset to those logits, rather than making an otherwise timestep-independent model time aware. Its token-dependent gate acts before the channel-mixing projection $W_o^s$, whereas $\alpha_s(t)$ scales the projected residual branch; these operations are generally not interchangeable. The distinction is placement and combination with content logits, not the use of SiLU~\citep{elfwing2017silu}. The two conditioners also differ in modulation type: adaLN-Zero applies an affine map with unbounded scale and shift $\gamma_s(t),\beta_s(t)\in\mathbb{R}^{D}$, whereas the sigmoid restricts each gate entry to $(0,1)$, so the time branch can only attenuate or pass an attention channel. This bounded, multiplicative form gives it an inductive bias toward selective suppression rather than arbitrary feature rescaling.

\subsection{Signal-to-noise view of the time bias}
\label{app:snr}
The content and time branches consume different inputs, which motivates an explicit time bias even though the content logits already depend on $t$ through $X_s(t)$ (Appendix~\ref{app:conditioning}). Write the content logits as $C_s(t)=X_s(t)W^s_{q,g}[\text{gate}]+b^s_{q,g}[\text{gate}]$. In flow/diffusion training~\citep{lipman2023flowmatching,ho2020ddpm}, the stream input carries a noisy latent $x_t=\alpha_t x_0+\sigma_t\epsilon$ with $\epsilon\sim\mathcal{N}(0,I)$, so the layer input decomposes as $X_s(t)=X_s^{\text{signal}}+X_s^{\text{noise}}$, where the first term carries semantic content and the second reflects residual noise. During the early, high-noise steps (large $t$, low SNR), the noise term dominates, $\|X_s^{\text{signal}}\|\ll\|X_s^{\text{noise}}\|$, and
\begin{equation}
C_s(t)\approx X_s^{\text{noise}}W^s_{q,g}[\text{gate}]+b^s_{q,g}[\text{gate}],
\label{eq:noisy-content-logits}
\end{equation}
so the token-dependent logits that should steer retention are themselves contaminated by noise precisely when reliable gating matters most.

The time bias avoids this pathway. Since $B_{\mathrm{ts},s}(t)=\operatorname{SiLU}(t_{\mathrm{emb}}(t))W^{\mathrm{time}}_s+b^{\mathrm{time}}_s$ is a deterministic function of the scalar timestep and never passes through the noisy latent, it supplies a noise-free, token-shared prior that is well defined at every noise level. Because $B_{\mathrm{ts},s}(t)$ is added before the sigmoid (Equation~\eqref{eq:gate}), it can set a stage-appropriate operating point for the gate when $C_s(t)$ is unreliable, then recede in relative influence as the SNR of $X_s(t)$ rises during late steps and the content logits recover their fidelity. This yields an implicit, learned curriculum in which the timestep prior governs retention early and cedes control to content selection later, without any explicit schedule.

\FloatBarrier

\section{Experimental configuration and evaluation protocol}
\label{app:setup}
\label{app:network-config}
\subsection{Architecture and training}
Table~\ref{tab:moe-network-config} brings together the architecture, gate settings, and parameter counts of all five evaluated MoE variants. Widening the Baseline's shared expert approximately matches TSGate's total parameter count, but the capacity allocation differs across variants. The fixed-weight interventions in Appendix~\ref{app:controls} all use the same trained MoE TSGate checkpoint.

\begin{table}[!htbp]
\caption{\textbf{Architecture and gate settings of the five MoE configurations.} Gated attention~\citep{qiu2025gatedattention} uses content-only gating; TSGate-L0 adds the time branch only at Layer 0. EW and HW denote element-wise and head-wise gating. Parameter counts are in millions (M) and include all saved streams; totals are rounded to 0.001M.}
\label{tab:moe-network-config}
\centering\small
\renewcommand{\arraystretch}{1.12}
\begin{tabular*}{\linewidth}{@{\extracolsep{\fill}}lrrrrr@{}}
\toprule
Configuration & Baseline & \shortstack{Gated\\attention} & TSGate & \shortstack{TSGate-\\L0} & \shortstack{TSGate-\\head} \\
\midrule
Transformer layers & 12 & 12 & 12 & 12 & 12 \\
Hidden width $D$ & 1,536 & 1,536 & 1,536 & 1,536 & 1,536 \\
Attention heads & 12 & 12 & 12 & 12 & 12 \\
Head dimension & 128 & 128 & 128 & 128 & 128 \\
Text input width & 3,584 & 3,584 & 3,584 & 3,584 & 3,584 \\
Routed experts $E$ & 8 & 8 & 8 & 8 & 8 \\
Active experts $k$ & 2 & 2 & 2 & 2 & 2 \\
Routed-expert FFN width & 1,024 & 1,024 & 1,024 & 1,024 & 1,024 \\
Shared experts & 1 & 1 & 1 & 1 & 1 \\
Shared-expert width & 5,120 & 2,048 & 2,048 & 2,048 & 2,048 \\
Text/audio FFN width & 1,536 & 1,536 & 1,536 & 1,536 & 1,536 \\
Content-gate granularity & -- & EW & EW & EW & HW \\
Content-gated layers & -- & 0--11 & 0--11 & 0--11 & 0--11 \\
Time-gate granularity & -- & -- & EW & EW & HW \\
Time-conditioned layers & -- & -- & 0--11 & 0 & 0--11 \\
Content-gate bias init. & -- & 4 & 4 & 4 & 4 \\
\midrule
Time-branch parameters (M) & 0.000 & 0.000 & 84.990 & 7.082 & 0.664 \\
Total parameters (M) & 1,250.706 & 1,165.826 & 1,250.816 & 1,172.909 & 1,082.164 \\
\bottomrule
\end{tabular*}
\end{table}

Table~\ref{tab:dense-network-config} lists the two dense checkpoints compared in Appendix~\ref{app:dense}. They share the 24-layer backbone dimensions; TSGate adds element-wise content gates in all 24 layers with biases initialized to 4 and a zero-initialized time projection in every layer. As in the MoE models, the saved checkpoints register image, text, and audio stream projections, so the dense time branch contributes $3\times24\times(1{,}536\times1{,}536+1{,}536)=169{,}979{,}904$ parameters, the same amount as its widened content-gate projections. No dense content-only gated-attention results are reported.

\begin{table}[!htbp]
\caption{\textbf{Dense backbone configurations for the ungated baseline and TSGate.} All 24 image FFNs are dense, so routed/shared-expert settings are inactive. Audio FFN width is a configured value, although generation uses image and text streams. Parameter counts are in millions (M) and include all saved streams; totals are rounded to 0.001M.}
\label{tab:dense-network-config}
\centering\small
\renewcommand{\arraystretch}{1.12}
\begin{tabular*}{\linewidth}{@{\extracolsep{\fill}}lrr@{}}
\toprule
Configuration & Baseline & TSGate \\
\midrule
Transformer layers & 24 & 24 \\
Hidden width $D$ & 1,536 & 1,536 \\
Attention heads & 12 & 12 \\
Head dimension & 128 & 128 \\
Text input width & 4,096 & 4,096 \\
Image FFN width & 3,072 & 3,072 \\
Text FFN width & 1,024 & 1,024 \\
Audio FFN width (configured) & 128 & 128 \\
Content-gate granularity & -- & Element-wise \\
Content-gated layers & -- & 0--23 \\
Time-gate granularity & -- & Element-wise \\
Time-conditioned layers & -- & 0--23 \\
Content-gate bias init. & -- & 4 \\
\midrule
Time-branch parameters (M) & 0.000 & 169.980 \\
Total parameters (M) & 1,167.000 & 1,506.960 \\
\bottomrule
\end{tabular*}
\end{table}

All configurations are trained on the curated 100M-image subset of LAION-5B~\citep{schuhmann2022laion5b} re-captioned with Qwen2.5-VL 72B~\citep{bai2025qwen25vl} into the structured-prompt format of Appendix~\ref{app:structured-prompt}, using 32 NVIDIA H200 GPUs. Training uses the AdamW optimizer~\citep{loshchilov2019adamw} with a learning rate of $1\times10^{-4}$, a weight decay of 0.01, and a maximum gradient norm of 1.0. The learning rate is warmed up over the first 1\% of training and then held constant. Training uses a text-drop ratio of 0.1 for classifier-free guidance~\citep{ho2022classifierfree} and a resolution of $320\times320$. SNR sampling follows a log-normal distribution with mean 0 and standard deviation 1 in log space. All models are trained on the same total of 100M images.

\subsection{Inference}
MoE inference uses CFG=5, $320\times320$ resolution, 50 sampling steps, timeshift=3, and seeds 42, 1234, 7777, and 8888. DPG-Bench~\citep{hu2024ella} scores the four-image grid for each source and averages its four member scores before averaging sources, on a 0--100 scale. GenEval~\citep{ghosh2023geneval} and T2I-CompBench~\citep{huang2023compbench} first average seeds within each source and then equally weight the six category means. \label{app:inputs}
Dense inference uses the same CFG, resolution, and sampling-step settings.

\section{Complete benchmark results}
\label{app:quality}
\label{sec:overall}
\label{sec:external}
This section collects the reported overall scores, category breakdowns, and paired confidence intervals for DPG-Bench~\citep{hu2024ella}, GenEval~\citep{ghosh2023geneval}, and T2I-CompBench~\citep{huang2023compbench}. The five variants are Baseline, Gated attention~\citep{qiu2025gatedattention} (content-only gating), TSGate, TSGate-L0 (first-layer time branch), and TSGate-head (head-wise gating) (Appendix~\ref{app:variants}). DPG scores use a 0--100 scale; GenEval and CompBench use a 0--1 scale. Higher scores indicate better performance on all benchmarks. Generation and aggregation protocols are specified in Appendix~\ref{app:setup}, and parameter settings are collected in Tables~\ref{tab:moe-network-config} and~\ref{tab:dense-network-config}. Differences use unrounded, source-paired scores after averaging scores over four seeds within each source. Reported 95\% confidence intervals (CIs) are pointwise percentile intervals~\citep{efron1993bootstrap} from 10,000 source-bootstrap resamples (seed 20260909), with within-category resampling for external benchmarks.

\subsection{DPG-Bench}
\label{app:dpg-results}
\label{app:dense}
\label{sec:dense}
\label{app:dpg-dense-categories}

Table~\ref{tab:dpg-moe} collects the raw and full-recaption scores for all five MoE configurations on the same 1,065 D-main sources. TSGate has the highest mean for both inputs, followed by the first-layer variant TSGate-L0. The TSGate$-$Gated-attention difference is positive for both inputs, whereas the paired TSGate-L0$-$TSGate intervals include zero. These are source-paired comparisons of the checkpoints in Table~\ref{tab:moe-network-config}, not estimates of training-seed variation.

\begin{table}[!htbp]
\caption{\textbf{MoE DPG-Bench scores and paired differences.} Raw and full-recaption prompts use the same 1,065 sources, with scores averaged over four seeds per source. Scores use a 0--100 scale. Bold and underlining mark the highest and second-highest means in each prompt condition; intervals are pointwise 95\% source-bootstrap confidence intervals (CIs).}
\label{tab:dpg-moe}
\centering\small
\renewcommand{\arraystretch}{1.1}
\begin{tabular*}{\linewidth}{@{\extracolsep{\fill}}lrrrr@{}}
\toprule
& \multicolumn{2}{c}{Raw} & \multicolumn{2}{c}{Full recaption} \\
\cmidrule(lr){2-3}\cmidrule(l){4-5}
Model & \multicolumn{2}{c}{DPG score} & \multicolumn{2}{c}{DPG score} \\
\midrule
Baseline & \multicolumn{2}{c}{55.483} & \multicolumn{2}{c}{78.637} \\
Gated attention & \multicolumn{2}{c}{58.976} & \multicolumn{2}{c}{78.327} \\
TSGate & \multicolumn{2}{c}{\textbf{60.749}} & \multicolumn{2}{c}{\textbf{79.355}} \\
TSGate-L0 & \multicolumn{2}{c}{\underline{60.424}} & \multicolumn{2}{c}{\underline{79.264}} \\
TSGate-head & \multicolumn{2}{c}{55.069} & \multicolumn{2}{c}{78.858} \\
\midrule
Contrast & $\Delta$ & 95\% CI & $\Delta$ & 95\% CI \\
TSGate $-$ Gated attention & +1.773 & $[0.709,2.870]$ & +1.028 & $[0.239,1.795]$ \\
TSGate-L0 $-$ TSGate & $-0.325$ & $[-1.291,0.615]$ & $-0.091$ & $[-0.727,0.530]$ \\
\bottomrule
\end{tabular*}
\end{table}

Both dense models use the architecture and inference settings in Appendix~\ref{app:setup}. Table~\ref{tab:dpg-dense-overall} compares the baseline and TSGate on common sources. The larger raw-prompt gain is consistent with the MoE pattern, while the full-recaption paired interval includes zero.

\begin{table}[!htbp]
\caption{\textbf{Dense DPG-Bench scores on common sources.} Scores use a 0--100 scale. $N$ counts paired sources; $\Delta$ is TSGate minus the baseline, computed before rounding. Intervals are pointwise 95\% source-bootstrap CIs for the paired differences.}
\label{tab:dpg-dense-overall}
\centering\small
\renewcommand{\arraystretch}{1.1}
\begin{tabular*}{\linewidth}{@{\extracolsep{\fill}}lrrrrr@{}}
\toprule
Input & $N$ & Baseline & TSGate & $\Delta$ & 95\% CI \\
\midrule
Raw & 1,064 & 41.619 & 64.159 & +22.540 & $[21.175,23.884]$ \\
Full recaption & 1,062 & 81.809 & 82.297 & +0.489 & $[-0.084,1.068]$ \\
\bottomrule
\end{tabular*}
\end{table}

Raw scores cover 1,064 sources per model. Full recaption covers 1,064 baseline and 1,062 TSGate sources; the saved-coverage means are 81.753 and 82.297 (difference 0.544). The paired table uses their 1,062 common sources; the category table retains the evaluator's original coverage and aggregation.

Table~\ref{tab:dpg-dense-categories} gives the dense backbone's five L1 categories (Global, Entity, Attribute, Relation, Other) and 13 L2 entries. Global repeats at L2 because it has no subdivisions. These category aggregates differ from source-level overall scores; MoE results are available only at the overall-score level.

\begin{table}[!htbp]
\caption{\textbf{Dense DPG-Bench category scores.} Raw and full-recaption scores (0--100) retain the evaluator's original aggregation and source coverage for first-level (L1) and second-level (L2) categories.}
\label{tab:dpg-dense-categories}
\centering\small
\renewcommand{\arraystretch}{1.08}
\begin{tabular*}{\linewidth}{@{\extracolsep{\fill}}lrrrr@{}}
\toprule
& \multicolumn{2}{c}{Raw} & \multicolumn{2}{c}{Full recaption} \\
\cmidrule(lr){2-3}\cmidrule(l){4-5}
Category & Baseline & TSGate & Baseline & TSGate \\
\midrule
\multicolumn{5}{l}{\textit{L1 categories}} \\
Global & 76.287 & 75.758 & 89.914 & 85.226 \\
Entity & 68.612 & 77.344 & 89.647 & 90.350 \\
Attribute & 65.061 & 78.597 & 89.015 & 89.309 \\
Relation & 75.489 & 85.862 & 89.395 & 91.396 \\
Other & 66.092 & 80.397 & 90.113 & 90.575 \\
\midrule
\multicolumn{5}{l}{\textit{L2 categories}} \\
Global / -- & 76.287 & 75.758 & 89.914 & 85.226 \\
Entity / whole & 62.459 & 75.391 & 90.496 & 90.068 \\
Entity / part & 71.914 & 75.591 & 86.853 & 81.919 \\
Entity / state & 77.866 & 79.223 & 90.194 & 92.351 \\
Attribute / color & 65.570 & 76.822 & 90.599 & 90.564 \\
Attribute / shape & 57.505 & 82.126 & 90.858 & 81.641 \\
Attribute / size & 61.628 & 76.580 & 85.992 & 85.515 \\
Attribute / texture & 65.952 & 82.421 & 87.500 & 90.249 \\
Attribute / other & 73.557 & 80.556 & 89.376 & 89.733 \\
Relation / spatial & 79.070 & 87.211 & 91.476 & 92.308 \\
Relation / non-spatial & 71.835 & 83.547 & 84.173 & 90.336 \\
Other / count & 59.878 & 86.484 & 90.650 & 84.167 \\
Other / text & 69.871 & 75.499 & 89.205 & 92.179 \\
\bottomrule
\end{tabular*}

\end{table}

\subsection{GenEval and T2I-CompBench: overall scores}
Table~\ref{tab:external} reports all five MoE variants, with intervals for absolute scores and paired differences. TSGate outperforms Gated attention on both benchmarks, with the highest GenEval mean and the second-highest CompBench mean, where TSGate-head leads. The CompBench TSGate$-$Baseline interval includes zero.

\begin{table}[!htbp]
\caption{\textbf{GenEval and T2I-CompBench macro scores and paired differences.} Scores use a 0--1 scale. The six categories are equally weighted after averaging scores over four seeds within each source; pointwise 95\% CIs use 10,000 within-category bootstrap resamples. All models use full recaption. Bold and underlining mark the highest and second-highest means for each benchmark.}
\label{tab:external}
\centering\small
\renewcommand{\arraystretch}{1.1}
\begin{tabular*}{\linewidth}{@{\extracolsep{\fill}}lrrrr@{}}
\toprule
& \multicolumn{2}{c}{GenEval} & \multicolumn{2}{c}{T2I-CompBench} \\
\cmidrule(lr){2-3}\cmidrule(l){4-5}
Model & Score & 95\% CI & Score & 95\% CI \\
\midrule
Baseline & 0.7861 & $[0.7583,0.8124]$ & 0.5224 & $[0.5057,0.5394]$ \\
Gated attention & 0.7862 & $[0.7598,0.8119]$ & 0.5155 & $[0.4984,0.5326]$ \\
TSGate & \textbf{0.8064} & $[0.7796,0.8318]$ & \underline{0.5236} & $[0.5070,0.5404]$ \\
TSGate-L0 & 0.7887 & $[0.7620,0.8147]$ & 0.5231 & $[0.5066,0.5398]$ \\
TSGate-head & \underline{0.7972} & $[0.7703,0.8231]$ & \textbf{0.5264} & $[0.5097,0.5432]$ \\
\midrule
Contrast & $\Delta$ & 95\% CI & $\Delta$ & 95\% CI \\
TSGate $-$ Gated attention & +0.0202 & $[0.0007,0.0401]$ & +0.0081 & $[0.0023,0.0141]$ \\
TSGate $-$ Baseline & +0.0203 & $[0.0025,0.0389]$ & +0.0012 & $[-0.0046,0.0067]$ \\
\bottomrule
\end{tabular*}
\end{table}

\subsection{GenEval: category breakdown}
\label{app:categories}
\label{app:geneval-categories}
Table~\ref{tab:geneval-categories} reports all six GenEval categories. TSGate's means exceed those of Gated attention in every category, with the largest differences in position and color attributes. The T2I-CompBench breakdown follows in Appendix~\ref{app:comp-categories}. Category means describe where the aggregate gains occur; paired uncertainty for the macro scores appears in Table~\ref{tab:external}.

\begin{table}[!htbp]
\caption{\textbf{GenEval category means.} Scores use a 0--1 scale. $N$ is the number of source prompts and $\Delta$ denotes TSGate$-$Gated attention, computed before rounding. Bold marks the highest mean in each row.}
\label{tab:geneval-categories}
\centering\small
\renewcommand{\arraystretch}{1.1}
\begin{tabular*}{\linewidth}{@{\extracolsep{\fill}}lrrrrrrr@{}}
\toprule
Category & $N$ & Baseline & \shortstack{Gated\\attention} & TSGate & \shortstack{TSGate-\\L0} & \shortstack{TSGate-\\head} & $\Delta$ \\
\midrule
Single object & 80 & 0.9500 & 0.9563 & \textbf{0.9625} & 0.9563 & 0.9437 & +0.0062 \\
Two objects & 99 & 0.8283 & 0.8308 & 0.8359 & \textbf{0.8485} & 0.8409 & +0.0051 \\
Counting & 80 & 0.6281 & 0.6375 & \textbf{0.6594} & 0.6250 & 0.6531 & +0.0219 \\
Colors & 94 & 0.8404 & 0.8404 & \textbf{0.8457} & 0.8351 & 0.8431 & +0.0053 \\
Position & 100 & 0.8175 & 0.7800 & 0.8250 & 0.7850 & \textbf{0.8325} & +0.0450 \\
Color attribute & 100 & 0.6525 & 0.6725 & \textbf{0.7100} & 0.6825 & 0.6700 & +0.0375 \\
\bottomrule
\end{tabular*}
\end{table}

\subsection{T2I-CompBench: category breakdown}
\label{app:comp-categories}
Table~\ref{tab:comp-categories} reports all six T2I-CompBench categories. The TSGate$-$Gated-attention difference is largest in spatial relations and color, while Gated attention has the highest complex-category mean and TSGate-head leads the overall macro score.

\begin{table}[!htbp]
\caption{\textbf{T2I-CompBench category means.} Scores use a 0--1 scale. $N$ is the number of source prompts and $\Delta$ denotes TSGate$-$Gated attention, computed before rounding. Bold marks the highest mean in each row.}
\label{tab:comp-categories}
\centering\small
\renewcommand{\arraystretch}{1.1}
\begin{tabular*}{\linewidth}{@{\extracolsep{\fill}}lrrrrrrr@{}}
\toprule
Category & $N$ & Baseline & \shortstack{Gated\\attention} & TSGate & \shortstack{TSGate-\\L0} & \shortstack{TSGate-\\head} & $\Delta$ \\
\midrule
Color & 150 & 0.8145 & 0.8092 & \textbf{0.8264} & 0.8217 & 0.8200 & +0.0172 \\
Shape & 150 & 0.5912 & 0.5876 & 0.5903 & 0.6006 & \textbf{0.6027} & +0.0027 \\
Texture & 150 & 0.6954 & 0.6802 & 0.6852 & \textbf{0.6998} & 0.6913 & +0.0050 \\
Spatial & 150 & 0.3409 & 0.3193 & 0.3444 & 0.3217 & \textbf{0.3480} & +0.0251 \\
Non-spatial & 150 & 0.3043 & 0.3033 & 0.3044 & 0.3038 & \textbf{0.3053} & +0.0011 \\
Complex & 150 & 0.3879 & \textbf{0.3932} & 0.3907 & 0.3913 & 0.3910 & $-0.0026$ \\
\bottomrule
\end{tabular*}
\end{table}

\section{Structured prompt format}
\label{app:structured-prompt}
Our text-to-image data use long English captions organized under structured headings. The examples below retain the five fields of their paired source captions: a summary, setting, subject details, visual style, and mood. Within each field, the description remains ordinary prose. The three image--caption pairs below are data examples.
\vspace{2cm}

\begingroup
\setlength{\fboxsep}{6pt}
\setlength{\fboxrule}{0.35pt}
\newcommand{\promptfield}[2]{{\color{blue!35!black}\bfseries\# #1}\par #2\par\vspace{2pt}}
\newcommand{\structuredexample}[4]{%
\par\noindent\fcolorbox{black!18}{black!2}{%
\begin{minipage}{\dimexpr\linewidth-2\fboxsep-2\fboxrule\relax}
\begin{minipage}[c]{0.29\linewidth}
\centering
\includegraphics[width=\linewidth,height=3.6cm,keepaspectratio]{#1}\par\smallskip
{\small\bfseries #2}\par
{\scriptsize\color{black!60}Data sample #3}
\end{minipage}\hfill
\begin{minipage}[c]{0.67\linewidth}
\raggedright\ttfamily\fontsize{7.8}{9.3}\selectfont
\setlength{\parskip}{0pt}#4
\end{minipage}
\end{minipage}}\par\vspace{8pt}}

\structuredexample{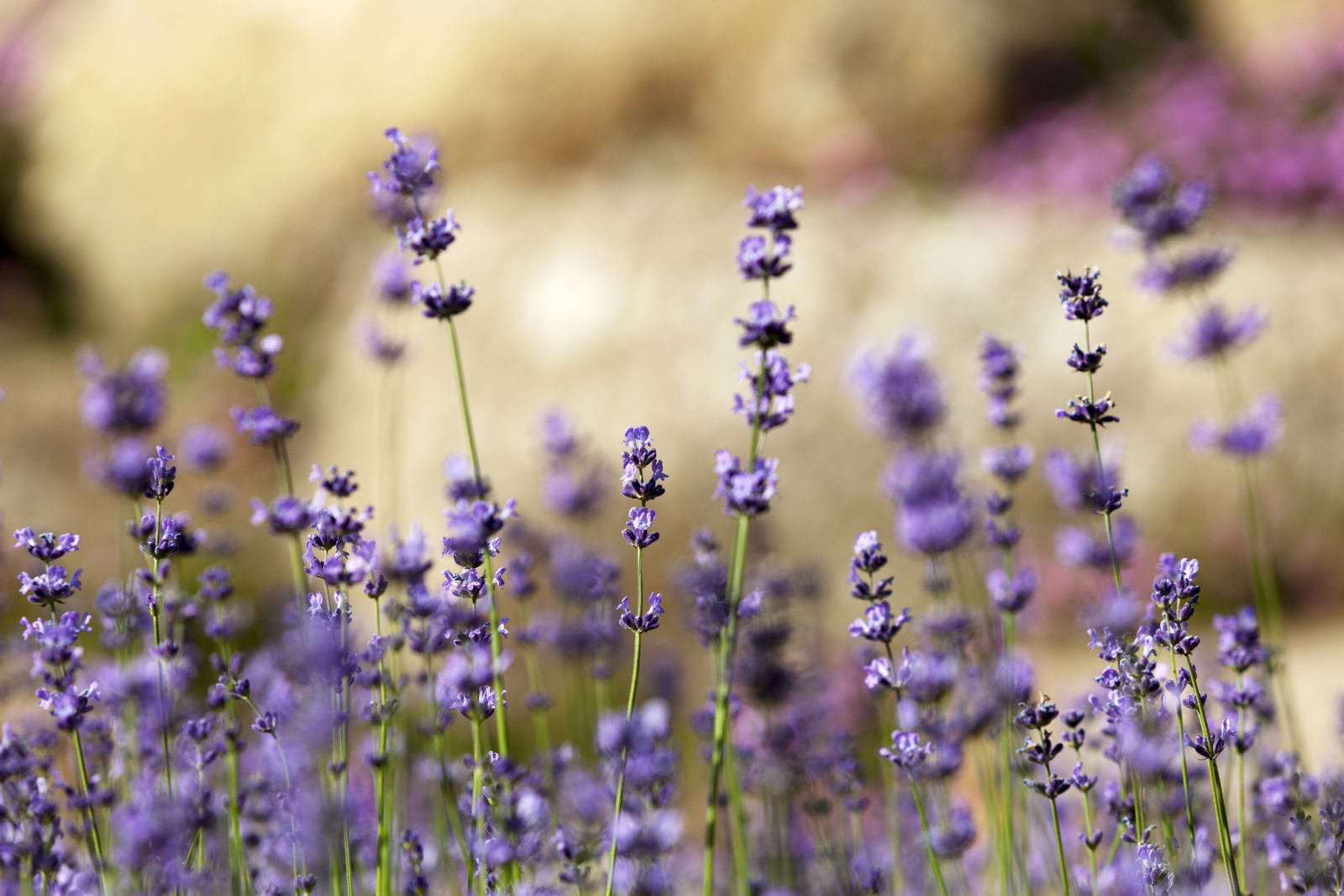}{(a) Lavender}{0439}{%
\promptfield{Summary}{A close-up of purple lavender blooms, with a few sharply focused stalks rising against a softly blurred, warm-toned background.}
\promptfield{Setting}{An outdoor lavender patch in soft, diffused daylight; the foreground stalks are crisp while the background dissolves into a smooth bokeh.}
\promptfield{Subject}{Slender light-green stems carry whorls of small violet and lilac flowers; a few stalks are in sharp focus while the rest fade into blur, densely packed near the base.}
\promptfield{Style}{Botanical macro close-up; a pronounced shallow depth of field isolates the flowers against a warm beige and sandy bokeh, with cool purples contrasting the earthy backdrop.}
\promptfield{Mood}{Quiet, delicate, and slightly dreamy; the soft light and gentle blur evoke a calm, intimate moment in nature.}}

\structuredexample{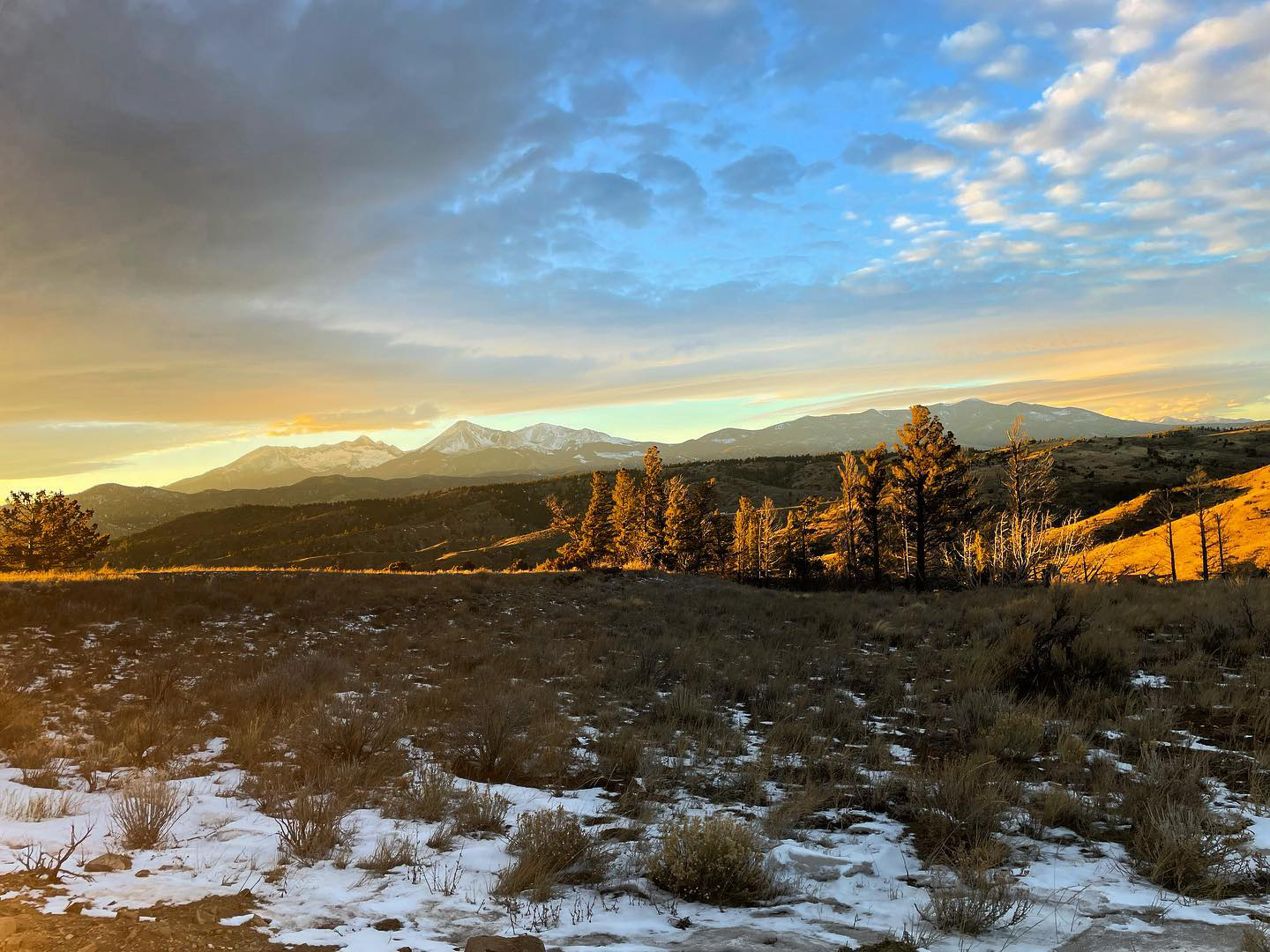}{(b) Mountain light}{0454}{%
\promptfield{Summary}{Golden low-sun light falls across rolling hills and a wooded ridge beneath a range of snow-capped mountains.}
\promptfield{Setting}{A mountainous landscape at sunrise or sunset; patchy snow and dry brush fill the foreground, a conifer-dotted ridge crosses the midground, and a broad, partly cloudy sky spans the horizon.}
\promptfield{Subject}{Sunlit conifers glow golden on the right of the ridge while the left falls into shadow; distant peaks show snow on their upper slopes against a gradient sky from warm yellow to deep blue.}
\promptfield{Style}{Wide landscape photograph with deep focus; a strong warm-cool contrast pairs golden highlights on the trees and hillside with cool blue sky and shadowed terrain.}
\promptfield{Mood}{Serene, expansive, and quietly dramatic; the interplay of warm light and cool shadow conveys peaceful isolation and the scale of nature.}}

\structuredexample{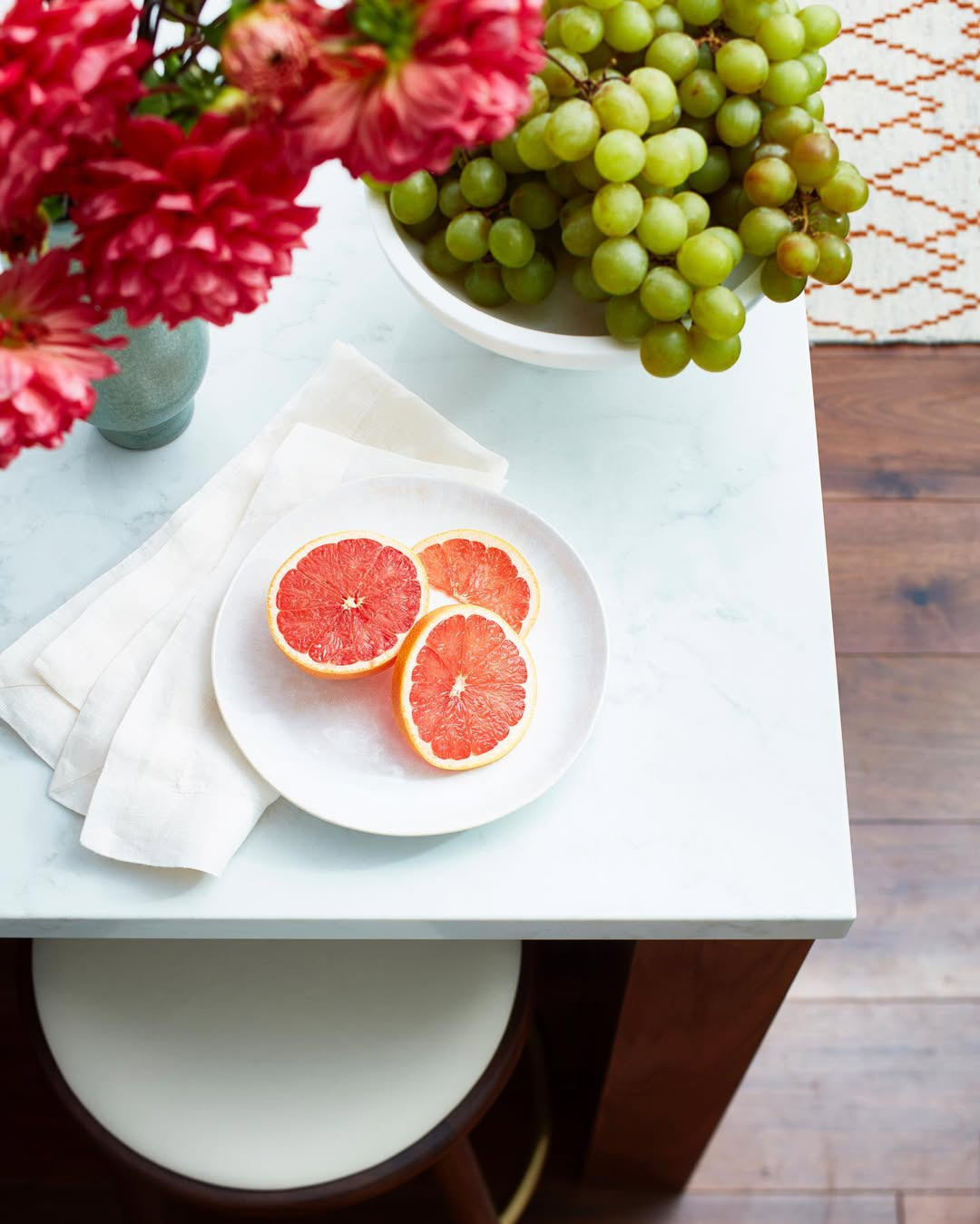}{(c) Fresh still life}{0861}{%
\promptfield{Summary}{A high-angle still life of pink grapefruit slices, green grapes, and deep red dahlias arranged on a white marble countertop.}
\promptfield{Setting}{A bright interior lit by soft window light from the upper left; the marble counter fills the frame, with a stool and wooden floor visible below and a patterned rug in the corner.}
\promptfield{Subject}{Three coral-pink grapefruit slices rest on a white plate beside a folded off-white linen napkin; a white bowl of glossy green grapes sits at the top right and crimson dahlias in a sage-green vase at the top left.}
\promptfield{Style}{High-angle lifestyle photograph; vivid coral, green, and crimson accents stand out against a high-key palette of white marble and pale linen, with soft shadows revealing each texture.}
\promptfield{Mood}{Fresh, calm, and inviting; the bright light and clean, casual arrangement suggest a relaxed, healthy morning.}}
\endgroup

\section{Additional paired generation comparisons}
\label{app:paired}
\label{sec:paired}
We provide additional qualitative comparisons across all three models, laid out two per row. In each panel the top row uses the full-recaption structured prompt (in domain) and the bottom row uses the raw prompt (out of domain); columns are Baseline, Gated attention, and TSGate. Within each example, all three models and both prompt forms use the same random seed.

\FloatBarrier

\begingroup
\setlength{\parskip}{0pt}
\newcommand{\paircase}[3]{%
\begin{minipage}[t]{0.48\linewidth}\centering
\includegraphics[width=\linewidth]{#1}\par\smallskip
\captionof{figure}{#2}\label{#3}
\end{minipage}}

\noindent
\paircase{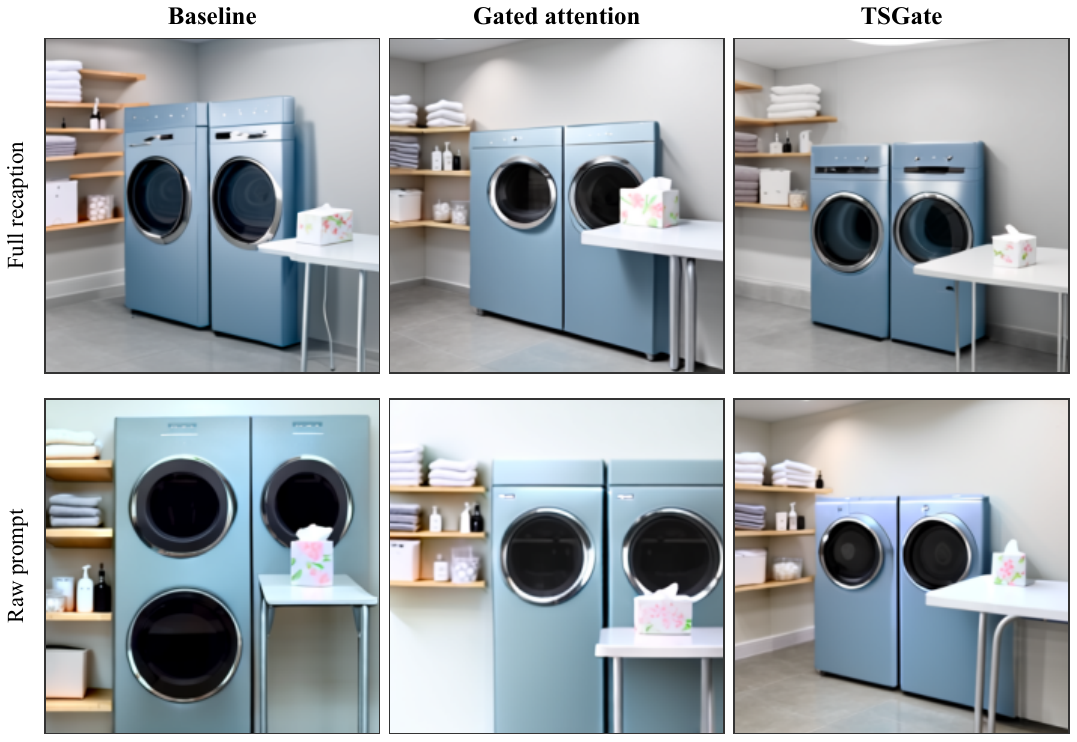}{Blue washing machines in a laundry room.}{fig:paired-image-laundry}\hfill
\paircase{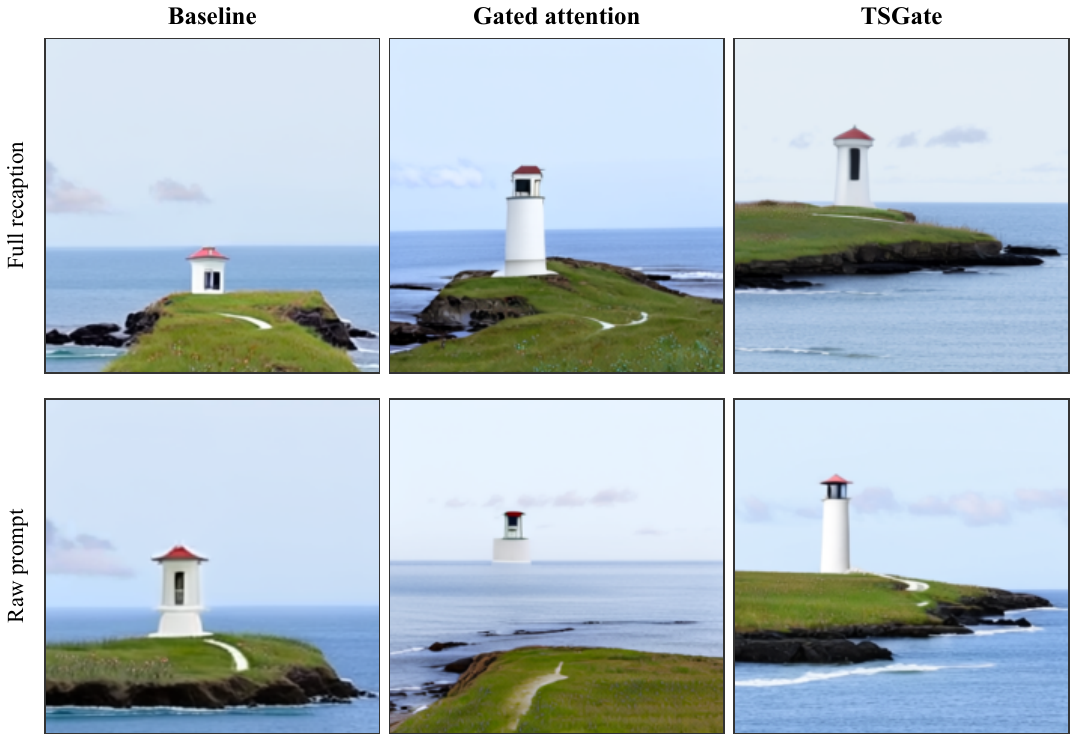}{A white lighthouse on a rocky coast.}{fig:paired-image-lighthouse}

\vspace{12pt}\noindent
\paircase{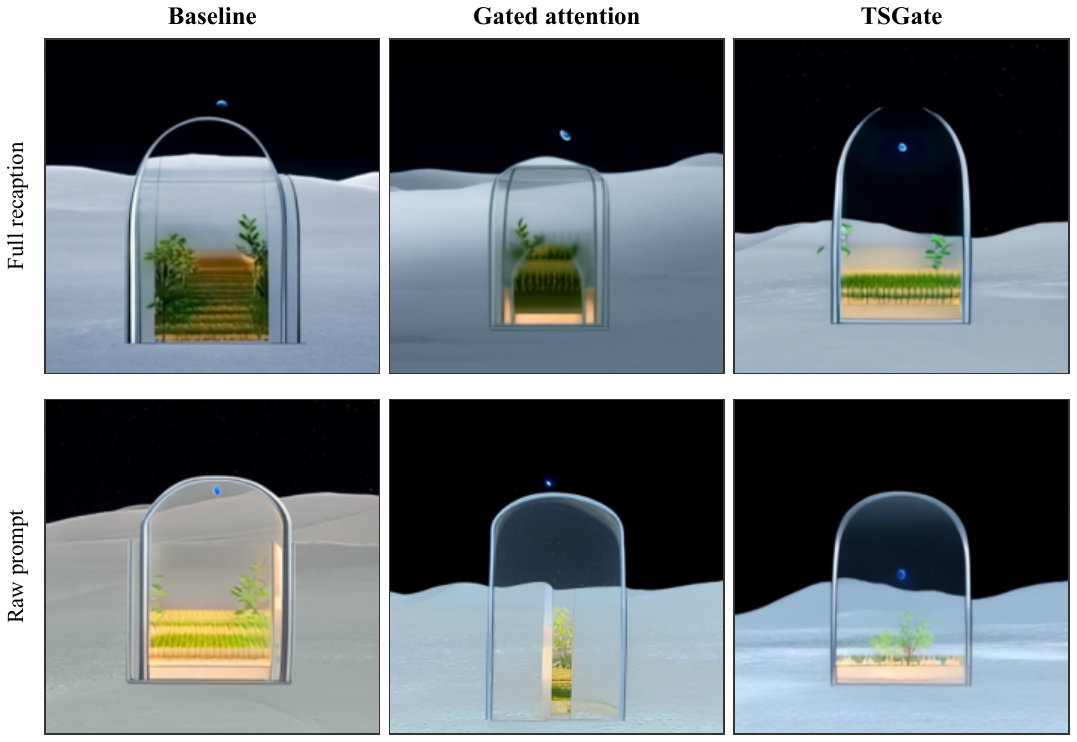}{A glass greenhouse on the Moon.}{fig:paired-image-lunar-greenhouse}\hfill
\paircase{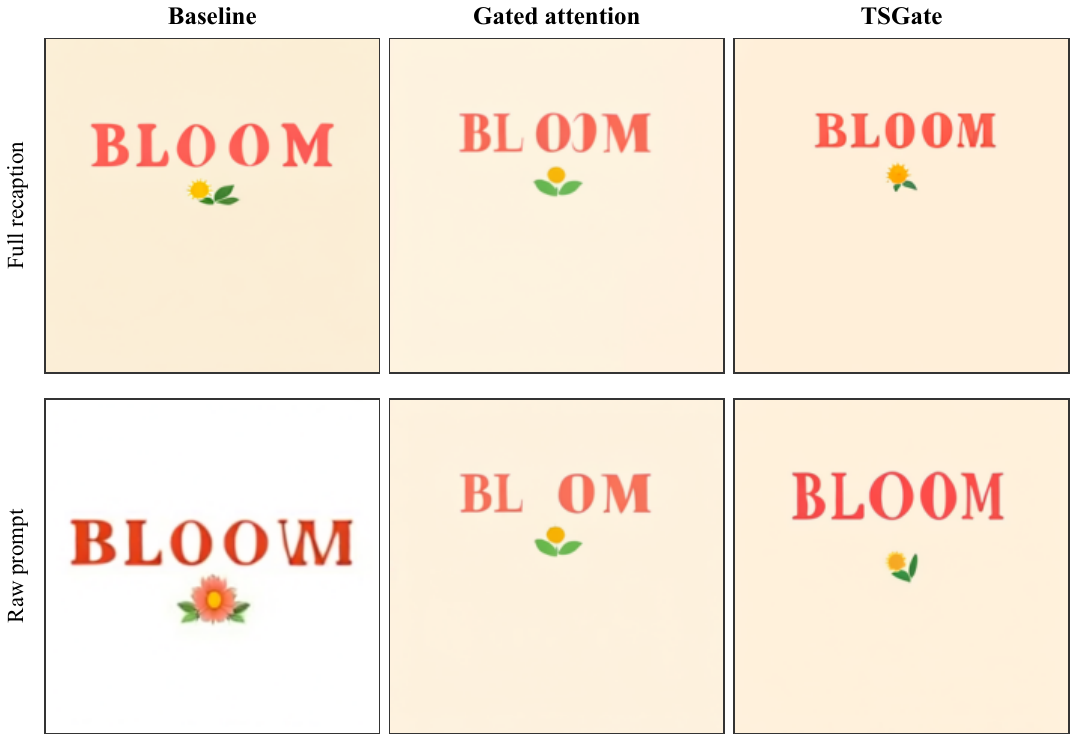}{A ``BLOOM'' poster with a small flower.}{fig:paired-image-bloom}

\vspace{12pt}\noindent
\paircase{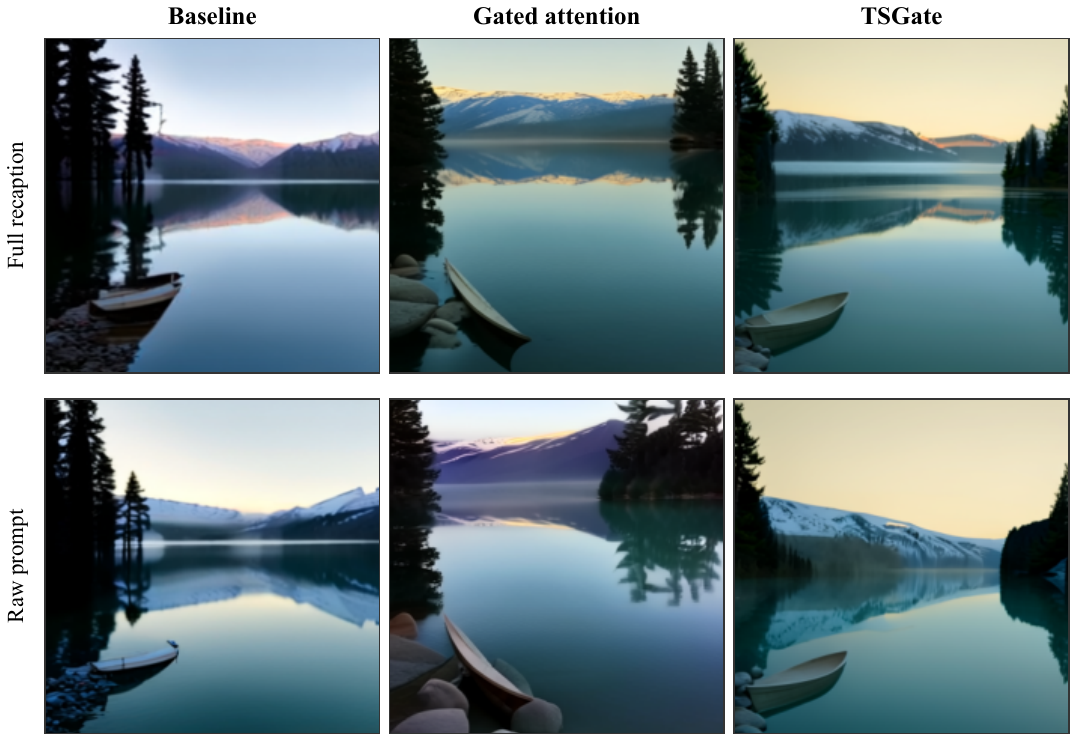}{A small boat beside a calm mountain lake.}{fig:paired-image-lakeside-boat}\hfill
\paircase{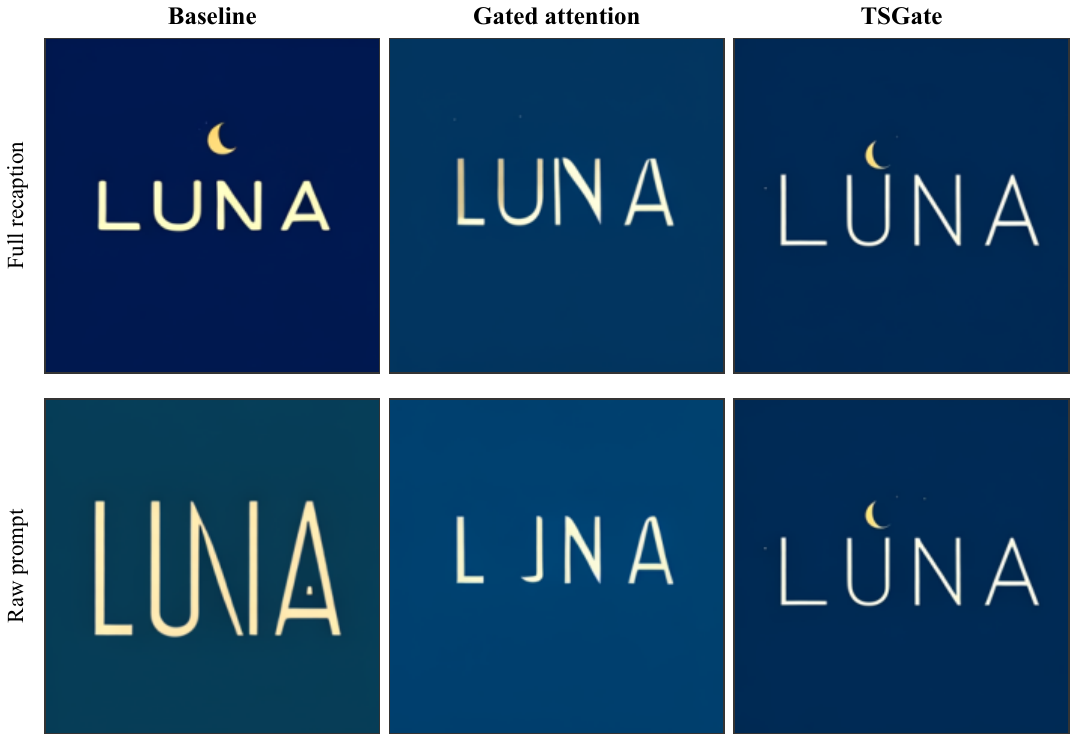}{A ``LUNA'' logo with a small crescent moon.}{fig:paired-image-luna}

\vspace{12pt}\noindent
\paircase{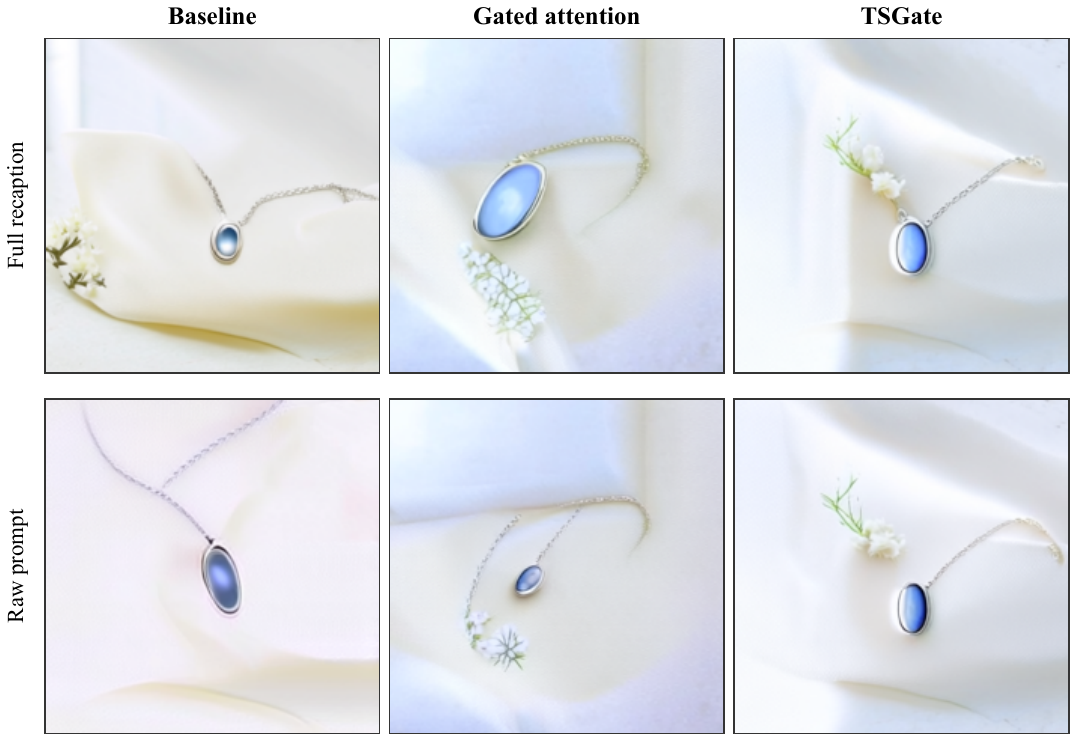}{A blue gemstone pendant on white fabric.}{fig:paired-image-blue-pendant}\hfill
\paircase{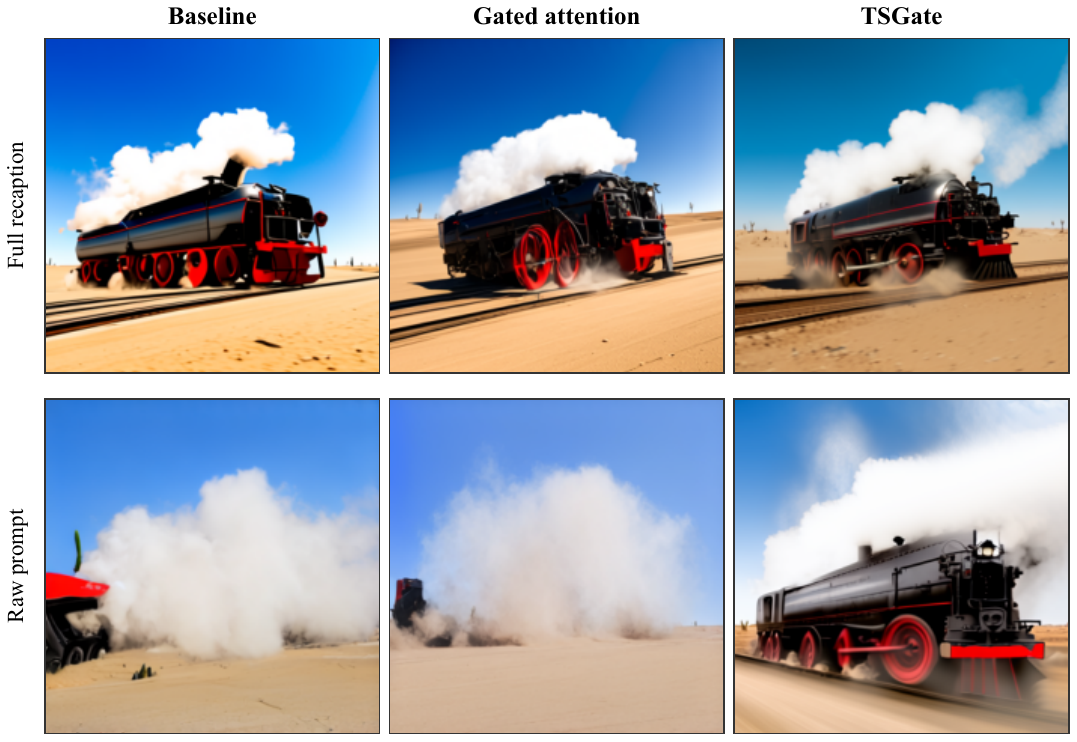}{A steam locomotive crossing a desert.}{fig:paired-image-steam-locomotive}
\endgroup

\FloatBarrier

\section{Stage-dependent text interaction and time-bias interventions}
\label{app:controls}
\label{sec:temporal}
\begin{figure}[!hbp]
\centering
\includegraphics[width=\linewidth]{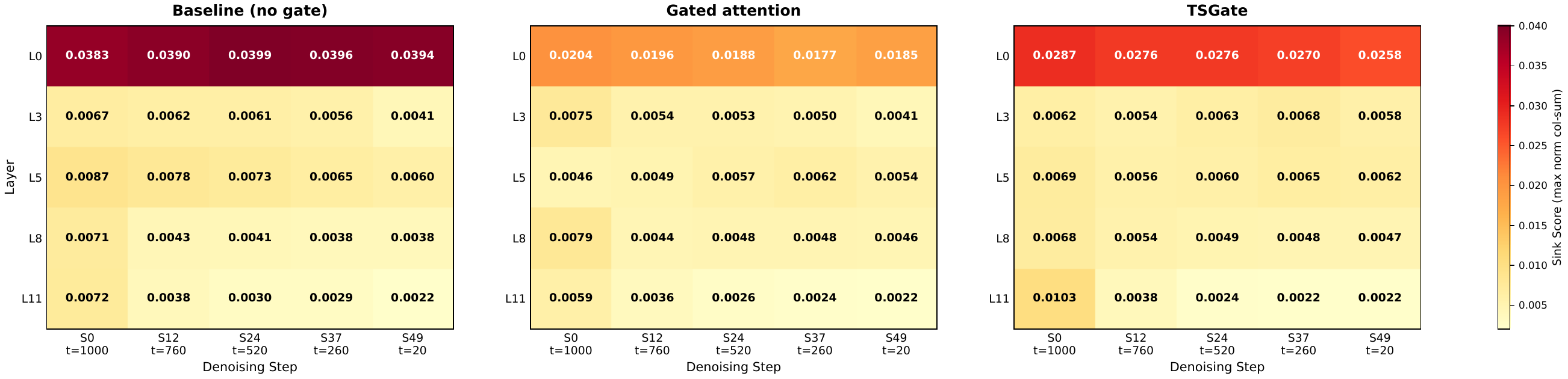}
\caption{\textbf{Attention sinks concentrate at Layer 0 across all three variants.} Heatmaps show prompt-averaged Sink Scores across layers and denoising steps. The Sink Score is the maximum normalized column-sum of the joint-attention map; darker cells indicate stronger sinks. Attention weights are from the conditional branch and measured before the output gate.}
\label{fig:attention}
\end{figure}

The strongest sink suppression does not yield the highest generation quality. Across layers and denoising steps, the Sink Score concentrates at Layer 0, where it follows Gated attention~\citep{qiu2025gatedattention} $<$ TSGate $<$ Baseline, whereas TSGate outperforms Gated attention on DPG~\citep{hu2024ella} (Figure~\ref{fig:attention}, Table~\ref{tab:dpg}). The Sink Score is the maximum normalized column-sum of the joint-attention map, i.e., the largest fraction of attention received by any single token; it is measured before the output gate, so it characterizes learned interactions and the hidden states propagated through the network rather than a same-operation renormalization by the gate.

\subsection{Fixed-weight interventions on time biases}
\label{sec:interventions}
To test the temporal role of TSGate in Section~\ref{sec:temporal}, we replace its explicit bias sequence while keeping the trained weights and backbone timestep conditioning fixed. The content branch remains active and responds to the hidden states of each intervened trajectory. For the complete schedule $t_0,\ldots,t_{K-1}$ with $K=50$, the bias used at step $k$ is
\begin{equation}
\begin{aligned}
\text{native}:&\quad\widetilde B(t_k)=B_{\mathrm{ts}}(t_k),\\
\text{reverse}:&\quad\widetilde B(t_k)=B_{\mathrm{ts}}(t_{K-1-k}),\\
\text{matched-mean}:&\quad\widetilde B(t_k)=\frac{1}{K}\sum_{j=0}^{K-1}B_{\mathrm{ts}}(t_j),\\
\text{zero}:&\quad\widetilde B(t_k)=0.
\end{aligned}
\label{eq:intervention}
\end{equation}
Layer and stream indices are omitted; each replacement is applied separately to every enabled layer, stream, and channel in Equation~\eqref{eq:gate}. Reverse preserves the bias values but reverses their order, testing alignment with the current stage. Matched-mean preserves the average bias and removes variation across steps. Zero measures reliance on the entire branch in the current model. Although zero recovers the content-only gate formula, it does not recover the independently trained Gated attention checkpoint: TSGate's weights are retained, and its content logits evolve along the intervened trajectory. Each intervention is paired with native inference using the same source prompt and generation seed.


Reverse assigns the bias from position $49-k$ to position $k$, matched-mean applies each channel's average over the 50-step schedule at every position, and zero removes the additive bias. With the TSGate checkpoint, generation seeds, and backbone timestep conditioning fixed, reverse reduces the raw DPG score by 2.614 points (95\% CI $[-4.281,-0.994]$), matched-mean changes it by $-0.929$ points with an interval including zero, and zero changes it by $-42.831$ points. The reverse result demonstrates that correspondence between the learned bias and denoising stage matters.

\begin{figure}[t]
\centering
\includegraphics[width=0.8\linewidth]{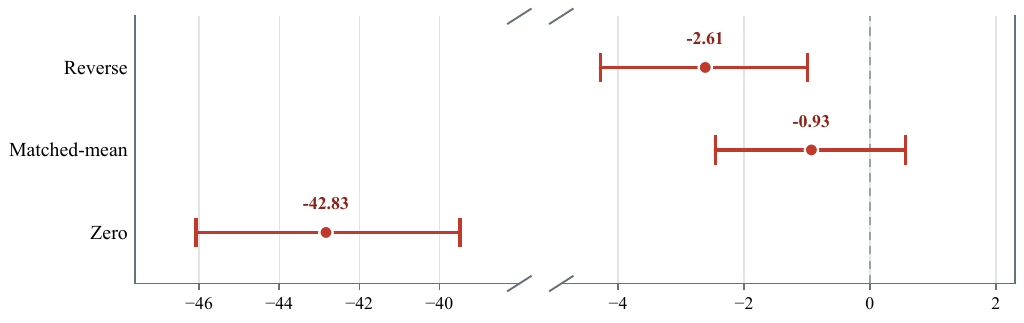}
\caption{\textbf{Effects of time-bias interventions with model weights fixed.} On 200 raw D-mech sources, points show mean DPG changes relative to native inference; whiskers show paired source-bootstrap 95\% CIs. The x-axis omits $(-38,-5)$ with equal unit spacing on the two segments; the dashed line marks zero.}
\label{fig:temporal}
\end{figure}

\subsection{Effects on raw and structured inputs}
Table~\ref{tab:all-interventions} reports nine contrasts on 200 paired sources. Reverse reduces the raw DPG score, with a confidence interval for the change that lies below zero, supporting the functional role of stage assignment. Zero produces a large reduction under both input constructions, showing reliance on the explicit branch in the trained model. Matched-mean preserves generation quality substantially better than zero.

\begin{table}[h]
\caption{\textbf{DPG changes under time-bias interventions.} $\Delta$ is the score change relative to native inference. The last three rows subtract the structured-prompt change from the raw-prompt change. All rows use 200 paired sources; intervals are pointwise 95\% source-bootstrap CIs.}
\label{tab:all-interventions}
\centering\small
\renewcommand{\arraystretch}{1.1}
\begin{tabular*}{\linewidth}{@{\extracolsep{\fill}}llrr@{}}
\toprule
Input / contrast & Intervention & DPG $\Delta$ & 95\% CI \\
\midrule
Raw & Reverse & $-2.614$ & $[-4.281,-0.994]$ \\
Raw & Matched-mean & $-0.929$ & $[-2.455,0.560]$ \\
Raw & Zero & $-42.831$ & $[-46.078,-39.485]$ \\
Structured & Reverse & $-0.919$ & $[-2.854,0.948]$ \\
Structured & Matched-mean & $-0.027$ & $[-1.625,1.619]$ \\
Structured & Zero & $-39.987$ & $[-43.457,-36.513]$ \\
Raw minus structured & Reverse & $-1.695$ & $[-4.329,0.989]$ \\
Raw minus structured & Matched-mean & $-0.902$ & $[-3.131,1.347]$ \\
Raw minus structured & Zero & $-2.844$ & $[-5.929,0.284]$ \\
\bottomrule
\end{tabular*}
\end{table}

\begin{figure}[t]
\centering
\includegraphics[width=\linewidth]{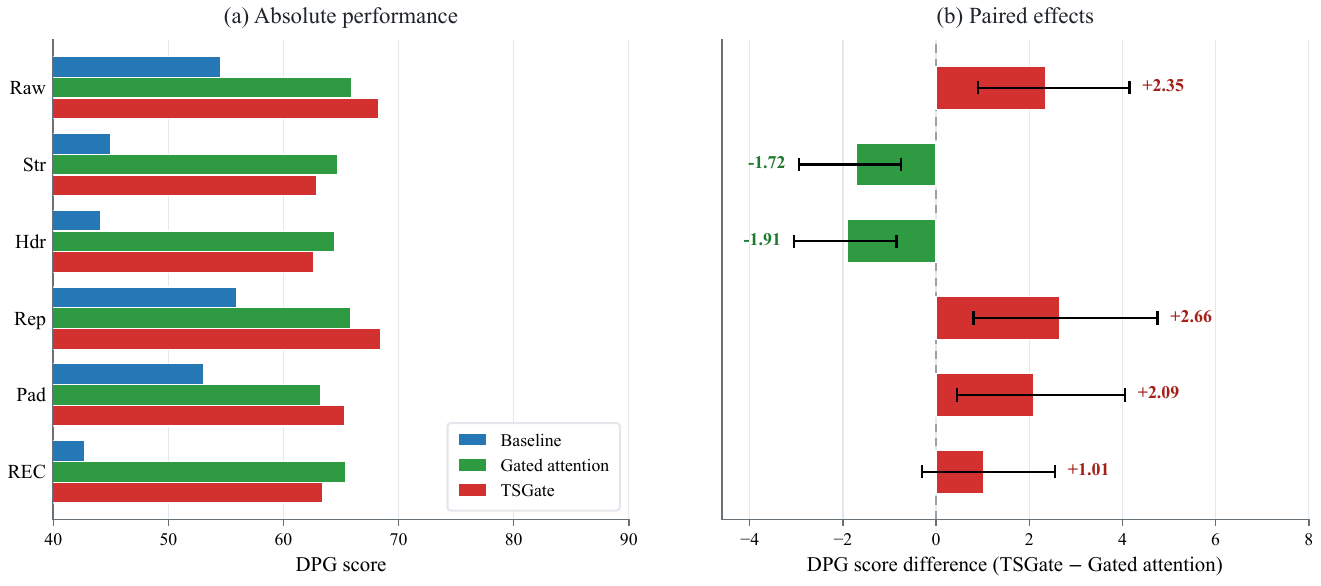}
\caption{\textbf{Generation quality across six prompt representations.} Both panels use the same 300 sources. \textbf{(a)} Mean DPG scores. \textbf{(b)} Paired differences (TSGate minus Gated attention) with pointwise source-bootstrap 95\% CIs; the dashed line marks zero. Raw: original; Str: structured; Hdr: header only; Rep: semantic repeat; Pad: neutral padding; REC: full recaption.}
\label{fig:representations}
\end{figure}

\section{Input representations and prompt robustness}
\label{app:representations}
\label{sec:representations}
We examine whether TSGate's gains depend on prompt rewriting or additional scene information, complementing work on prompt optimization. For this experiment, we sample 300 source prompts from DPG-Bench~\citep{hu2024ella} and evaluate Baseline, Gated attention~\citep{qiu2025gatedattention}, and TSGate on six input constructions, using the same sources and original scoring questions throughout.

Raw retains the original prompt. Semantic repeat duplicates that text without adding scene information. Neutral padding appends low-information text. Structured reorganizes the original description using headings while preserving its scene information; header only adds fixed headings. The structured condition here and in Table~\ref{tab:all-interventions} is distinct from the full-recaption structured prompts in the main text. These constructions probe sensitivity to repetition, appended text, organization, and rewriting; they are distinct input categories rather than successive levels of degradation.

Figure~\ref{fig:representations} shows that TSGate achieves higher mean DPG scores than Baseline under all six constructions. Relative to Gated attention, its gains are clearest on raw and semantic repeat, where the paired 95\% confidence intervals lie above zero. The intervals for the other four constructions include zero, and the mean differences for structured and header only are negative. These results show that TSGate can improve over content-only gating without recaptioning or additional scene information.

\end{document}